\documentclass[journal]{IEEEtran}
\usepackage{moreverb,url}
\usepackage[colorlinks,bookmarksopen,bookmarksnumbered,citecolor=blue,urlcolor=red]{hyperref}
\usepackage{graphics} 
\usepackage{epsfig} 
\usepackage{times} 
\usepackage{amsmath} 
\usepackage{amssymb}  
\usepackage[ruled,vlined,linesnumbered]{algorithm2e}
\usepackage{float}
\usepackage{algpseudocode}
\usepackage{booktabs}
\usepackage{graphicx}
\usepackage[table,xcdraw,dvipsnames]{xcolor}
\usepackage{multirow}
\usepackage{xcolor,pifont}
\usepackage{flushend}
\usepackage{siunitx}

\usepackage[numbers]{natbib}
\newcommand*\colourcheck[1]{%
  \expandafter\newcommand\csname #1check\endcsname{\textcolor{#1}{\ding{52}}}%
}
\colourcheck{green}

\newcommand*\colourcross[1]{%
  \expandafter\newcommand\csname #1cross\endcsname{\textcolor{#1}{\ding{55}}}%
}
\colourcross{red}

\usepackage{amsmath, amssymb}

\usepackage{xifthen}
\usepackage{color}

\newcommand{\ba}{\begin{eqnarray}}
\newcommand{\ea}{\end{eqnarray}}

\newcommand{\R}{\mathbb{R}}

\newcommand{\presup}[1]{\,{}^{\scriptscriptstyle #1}\!}

\newcommand{\pose}[1][ZZZZ]{\ifthenelse{\equal{#1}{ZZZZ}}{}{\presup{#1}}{\mathbf{\xi}}}
\newcommand{\estpose}[1][ZZZZ]{\ifthenelse{\equal{#1}{ZZZZ}}{}{\presup{#1}}{\mathbf{\hat{\xi}}}}
\newcommand{\hpose}[1][ZZZZ]{\ifthenelse{\equal{#1}{ZZZZ}}{}{\presup{#1}}{\hat{\mathbf{\xi}}}}
\newcommand{\posedot}[1][ZZZZ]{\ifthenelse{\equal{#1}{ZZZZ}}{}{\presup{#1}}{\mathbf{\nu}}}
\newcommand{\q}[1][ZZZZ]{\ifthenelse{\equal{#1}{ZZZZ}}{}{\presup{#1}}{\mathring{q}}}
\DeclareMathAlphabet{\mathitbf}{OML}{cmm}{b}{it}
\newcommand{\twist}[2][ZZZZ]{\ifthenelse{\equal{#1}{ZZZZ}}{}{\presup{#1}}{\mathcal{S}}}
\renewcommand{\vec}[2][ZZZZ]{\ifthenelse{\equal{#1}{ZZZZ}}{}{\presup{#1}}{\mathitbf{#2}}}
\newcommand{\hvec}[2][ZZZZ]{\ifthenelse{\equal{#1}{ZZZZ}}{}{\presup{#1}}{\tilde{\vec{#2}}}}
\newcommand{\evec}[2][ZZZZ]{\ifthenelse{\equal{#1}{ZZZZ}}{}{\presup{#1}}{\hat{\vec{#2}}}}
\newcommand{\bvec}[2][ZZZZ]{\ifthenelse{\equal{#1}{ZZZZ}}{}{\presup{#1}}{\bar{\vec{#2}}}}
\newcommand{\dvec}[2][ZZZZ]{\ifthenelse{\equal{#1}{ZZZZ}}{}{\presup{#1}}{\dot{\vec{#2}}}}
\newcommand{\ddvec}[2][ZZZZ]{\ifthenelse{\equal{#1}{ZZZZ}}{}{\presup{#1}}{\ddot{\vec{#2}}}}

\newcommand{\mat}[2][ZZZZ]{\ifthenelse{\equal{#1}{ZZZZ}}{}{\presup{#1}\,}{{\boldsymbol #2}}}
\newcommand{\dmat}[2][ZZZZ]{\ifthenelse{\equal{#1}{ZZZZ}}{}{\presup{#1}\,}{{\dot{\boldsymbol #2}}}}
\newcommand{\emat}[2][ZZZZ]{\ifthenelse{\equal{#1}{ZZZZ}}{}{\presup{#1}\,}{\hat{\boldsymbol#2}}}
\newcommand{\matfn}[3][ZZZZ]{\ifthenelse{\equal{#1}{ZZZZ}}{}{\presup{#1}}{{\mat{#2}}\left(#3\right)}}
\newcommand{\Rt}[2][ZZZZ]{\ifthenelse{\equal{#1}{ZZZZ}}{}{\presup{#1}}{{\bf R}\left(#2\right)}}

\newcommand{\point}[2][ZZZZ]{\ifthenelse{\equal{#1}{ZZZZ}}{}{\presup{#1}}{\mathbf{\mathrm{#2}}}}

\newfont{\School}{pncr}
\newfont{\eightTR}{pncr at 8pt}

\usepackage{color}
\usepackage{fancyvrb}
\fvset{formatcom=\color{blue},fontseries=c,fontfamily=courier,xleftmargin=4mm,commentchar=!}
\DefineVerbatimEnvironment{Code}{Verbatim}{formatcom=\color{blue},fontseries=c,fontfamily=courier,fontsize=\footnotesize,xleftmargin=4mm,commentchar=!}
\DefineVerbatimEnvironment{CodeSmall}{Verbatim}{formatcom=\color{blue},fontseries=c,fontfamily=courier,fontsize=\scriptsize,xleftmargin=1mm,commentchar=!}
\DefineVerbatimEnvironment{CodeNum}{Verbatim}{numbers=left,numbersep=4pt,formatcom=\color{blue},fontseries=c,fontfamily=courier,fontsize=\footnotesize,xleftmargin=4mm}

\newcommand{\model}[1]{\index{code}{#1@\textit{#1}}\ifthenelse{\boolean{draft}}{{\color{green}\Verb+#1+}}{\Verb+#1+}}
\newcommand{\block}[1]{\ifthenelse{\boolean{draft}}{{\color{green}\Verb+#1+}}{\textsf{#1}}}
\newcommand{\func}[2][ZZZZ]{\ifthenelse{\equal{#1}{ZZZZ}}{\index{code}{#2}}{\index{code}{#1}}\ifthenelse{\boolean{draft}}{{\color{green}\Verb+#2+}}{\Verb+#2+}}
\newcommand{\methodb}[2]{\index{code}{#1@\textbf{#1}!.#2}\ifthenelse{\boolean{draft}}{{\color{magenta}\Verb+#1.#2+}}{\Verb+#1.#2+}}
\newcommand{\method}[2]{\index{code}{#1@\textbf{#1}!.#2}\ifthenelse{\boolean{draft}}{{\color{magenta}\Verb+#2+}}{\Verb+#2+}}
\newcommand{\class}[1]{\index{code}{#1@\textbf{#1}}\ifthenelse{\boolean{draft}}{{\color{cyan}\Verb+#1+}}{\Verb+#1+}}
\newcommand{\property}[1]{\index{property}{#1}\ifthenelse{\boolean{draft}}{{\color{cyan}\Verb+#1+}}{\Verb+#1+}}

\newcommand{\SE}[1]{\ensuremath{\mathrm{{\bf SE}(#1)}}}

\newcommand\BibTeX{{\rmfamily B\kern-.05em \textsc{i\kern-.025em b}\kern-.08em
T\kern-.1667em\lower.7ex\hbox{E}\kern-.125emX}}

\let\oldnl\nl
\newcommand{\nonl}{\renewcommand{\nl}{\let\nl\oldnl}}

\begin{document}
\title{Bayesian Controller Fusion: Leveraging Control Priors In Deep Reinforcement Learning for Robotics}
%
%
%

\author{ \vspace{0.3cm} Krishan~Rana*, Vibhavari~Dasagi, Jesse Haviland, Ben Talbot, Michael Milford and Niko S\"underhauf\\ \vspace{0.3cm}
QUT Centre for Robotics
\thanks{\textbf{*Corresponding Author}:\\ Krishan Rana,
QUT Centre for Robotics, Queensland University of Technology, Brisbane, Australia. \texttt{ranak@qut.edu.au}}}

\maketitle

\begin{abstract}
We present Bayesian Controller Fusion (BCF): a hybrid control strategy that combines the strengths of traditional hand-crafted controllers and model-free deep reinforcement learning (RL). BCF thrives in the robotics domain, where reliable but suboptimal control priors exist for many tasks, but RL from scratch remains unsafe and data-inefficient. By fusing uncertainty-aware distributional outputs from each system, BCF arbitrates control between them, exploiting their respective strengths. We study BCF on two real-world robotics tasks involving navigation in a vast and long-horizon environment, and a complex reaching task that involves manipulability maximisation. For both these domains, simple handcrafted controllers exist that can solve the task at hand in a risk-averse manner but do not necessarily exhibit the optimal solution given limitations in analytical modelling, controller miscalibration and task variation. As exploration is naturally guided by the prior in the early stages of training, BCF accelerates learning, while substantially improving beyond the performance of the control prior, as the policy gains more experience. More importantly, given the risk-aversity of the control prior, BCF ensures safe exploration \emph{and} deployment, where the control prior naturally dominates the action distribution in states unknown to the policy. We additionally show BCF's applicability to the zero-shot sim-to-real setting and its ability to deal with out-of-distribution states in the real world. BCF is a promising approach towards combining the complementary strengths of deep RL and traditional robotic control, surpassing what either can achieve independently. The code and supplementary video material are made publicly available at \url{https://krishanrana.github.io/bcf}.

\end{abstract}


\maketitle

\section{Introduction}

\begin{figure}[t]
  \centering
  \includegraphics[width=0.45\textwidth]{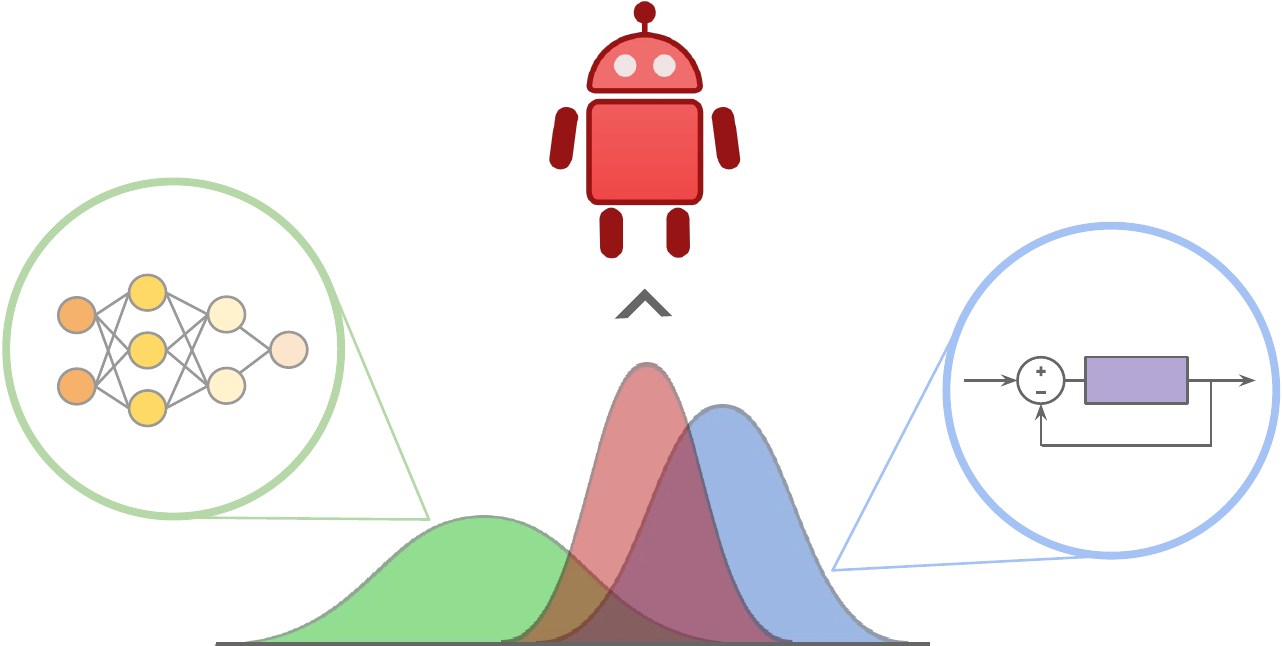}
  \caption{Bayesian Controller Fusion: We learn a compositional policy (red) for robotic agents that combines an uncertainty-aware deep RL policy (green) and a classical handcrafted controller (blue). Utilising this compositional policy to govern exploration allows for accelerated learning towards an optimal policy and safe behaviours in unknown states. It additionally allows for the reliable sim-to-real transfer of RL policies.}
  \label{bcf_simple}
\end{figure}

As the adoption of autonomous robotic systems increases around us, there is a need for the controllers driving them to exhibit the level of sophistication required to operate in our everyday unstructured environments. Recent advances in reinforcement learning (RL) coupled with deep neural networks as function approximators, have shown impressive results across a range of complex control tasks in robotics including dexterous in-hand manipulation \cite{dexterous}, quadrupedal locomotion \cite{haarnoja2018learning}, and targeted throwing \cite{throwing}. 

Nevertheless, the widespread adoption of deep RL for robot control is bottle-necked by two key factors: \textit{sample efficiency} and \textit{safety} \cite{Ibarz_2021}. Learning these behaviours requires large amounts of potentially unsafe interaction with the environment and the deployment of these systems in the real world comes with little to no performance guarantees. The latter can be attributed to the black-box nature of neural networks, while the sample complexity is related to the fact that RL agents tend to randomly search the solution space. This is further exacerbated by the sparse, long-horizon reward setting.

A general avenue to addressing the sample complexity in RL is the deliberate use of inductive bias or prior knowledge to aid the exploratory process. This includes reward-shaping \cite{schoettler2019deep, andrychowicz2018hindsight, rew_shaping_theoretical}, curriculum learning \cite{bengio2009curriculum}, learning from demonstrations \cite{hester2018deep, vecerik2017leveraging}, and the use of behavioural priors \cite{johannink2018residual, jeong2020learning, rana_mcf, lee2020guided}. The incorporation of prior knowledge in the form of behavioural priors has been gaining increasing traction in recent years. These priors include learned policies or hand-crafted controllers that capture the core capabilities for solving a task. They, however, are not necessarily the optimal solution to solving the task at hand. We refer to this class of methods as \textit{Reinforcement Learning from Behavioural Priors} (RLBP). RLBP approaches can directly query an action from the prior at any given state. This allows for a diverse range of mechanisms for introducing inductive bias during training, including regularisation \cite{tirumala2019exploiting_, galashov2019information_, cheng2019control}, exploration bias \cite{jeong2020learning, rana_mcf, cheng2019control} and residual learning \cite{johannink2018residual, silver2018residual}.

We focus on the robotics setting, where decades of research have yielded numerous behavioural priors in the form of hand-crafted controllers and algorithmic approaches for the vast majority of real-world physical systems (from mobile robots to humanoids) and tasks \cite{siciliano2016springer}. These include classical feedback controllers \cite{nise2020control}, trajectory generators \cite{ijspeert2008central} and behaviour trees \cite{colledanchise2016behavior}. While the explicit analytic nature of these controllers comes with certain performance guarantees including safety and robustness, they can be highly suboptimal with respect to variations in the task and tend to require great effort in system modelling and controller tuning to achieve higher levels of performance.

In this work, we present Bayesian Controller Fusion (BCF), a novel RLBP approach that closely couples hand-crafted robot controllers within the RL framework for accelerated learning and safety awareness during both training and deployment. BCF enables RL agents to learn substantially better behaviours than the underlying hand-crafted controllers while exploiting their risk-aversity for safe behaviours in out-of-distribution states, a limiting factor for neural network-based policies. 

In order to exploit the respective strengths of each of these systems, we draw inspiration from the dual-process theory  of decision-making, seen in the human brain. This theory suggests that the brain leverages multiple neural controllers to govern behaviours based on their confidence in a given scenario \cite{daw2005uncertainty_, dayan2008decision, hassabis2017neuroscience}. As opposed to leveraging just a single controller, this hybrid control strategy allows humans to exploit the strengths of each system when necessary in order to successfully complete a task. BCF follows this strategy, learning a compositional policy that interpolates between the behaviours specified by uncertainty-aware stochastic outputs generated by an RL policy and a hand-crafted controller as shown in Figure \ref{bcf_simple}. Our Bayesian formulation allows the system exhibiting the least uncertainty to dominate control.  This has important implications both during training and deployment. In states of high policy uncertainty, BCF biases the composite action distribution heavily towards the risk-averse prior, reducing the chances of catastrophic failure. In more confident states, it exploits the optimal behaviours discovered by the policy. We highlight our key contributions below:

\begin{itemize}
    \item A novel uncertainty-aware training strategy that accelerates learning in sparse reward, long-horizon task settings while allowing for safe exploration in unknown environments. 
    \item The ability to leverage a suboptimal controller to aid learning without inhibiting the final policy from identifying the optimal behaviours.
    \item A novel deployment strategy for robot control that combines the strengths of classical controllers and learned policies, allowing for reliable performance even in out-of-distribution states.
    \item An evaluation of our deployment strategy to transfer a simulation-trained policy directly to the real-world, for two different free-space motion robotics tasks, and its ability to act reliably in out-of-distribution states.
\end{itemize}

This paper is structured as follows. In Section \ref{sec:background}, we provide an overview of the RL concepts and algorithms used in this work and introduce the notion of control priors for robotics. Section \ref{sec:related_work} discusses related work and based on their limitations we formulate the problem in Section \ref{sec:pf}. In Section \ref{sec:bcf} we describe our algorithm, the derivation of the relevant components and BCFs applicability to both exploration and sim-to-real transfer. Section \ref{sec:experiments_setup} describes our experimental setup and in Section \ref{sec:training_performance} and \ref{sec:evaluation_deployed} we provide both qualitative and quantitative results for training and sim-to-real deployment of our system on a physical robot. In Section \ref{sec:limitations}, we highlight the limitations of our strategy and avenues for future work, before concluding with Section \ref{sec:conclusion}.

\section{Background}
\label{sec:background}
In this section, we provide an overview of two broad areas for robot control which we aim to combine: learned controllers based on the deep reinforcement learning framework and classical hand-crafted controllers. 

\subsection{Reinforcement Learning}

We consider the reinforcement learning framework in which an agent learns an optimal policy for a given task through environment interaction in order to solve a Markov Decision Process \cite{sutton1998introduction}. A Markov Decision Process (MDP) is described by a tuple $(\mathcal{S}, \mathcal{A}, \pi, \mathcal{P},r,\gamma, T)$ where: $\mathcal{S}$ is the set of all possible states and $\mathcal{A}$ is the set of all possible actions. $\pi : \mathcal{S} \times \mathcal{A}\mapsto[0,1]$ is a stochastic policy mapping state-action pairs $(s_t,a_t)$ to the probability of choosing an action $a_t\in\mathcal{A}$ when in state $s_t\in\mathcal{S}$ at timestep $t$. $\mathcal{P} : \mathcal{S}\times\mathcal{A}\times\mathcal{S} \mapsto[0,1]$ is a state transition function mapping tuples $(s_{t}, a_{t}, s_{t+1})$ to the probability of arriving in state $s_{t+1}$ after taking action $a_{t}$ at state $s_{t}$. $r: \mathcal{S} \times \mathcal{A} \mapsto[\mathbb{R}]$ is the reward function that assigns a scalar reward value to each state-action pair. $\gamma \in (0,1)$ is a discount factor and $T\in\mathbb{N}$ is the time horizon.

Using an MDP, we can generate a sequence of states and actions as follows. Given an initial state $s_{0}$, iteratively select the next action $a_{t} \sim \pi(s_{t})$ and evolve the state by sampling $s_{t+1}\sim\mathcal{P}(s_{t},a_{t})$. This generates a sequence of states and actions $\tau = \{s_{t}, a_{t}, ..., a_{T-1}, s_{T-1} \}$, called a trajectory. A reinforcement learning algorithm seeks to obtain an optimal policy, $\pi_{\theta}(a\mid s)$, by optimising over the policy parameters, $\theta$, such that the expected sum of discounted rewards across a trajectory, $J(\theta)$, is maximised:
\begin{align}
    \label{rl_obj}
    J(\theta) = \mathbb{E}_{\tau \sim \pi_{\theta}} \left[\sum_{t=0}^{T}\gamma^{t}r\left(s_{t}, a_{t}\right)\right] ,
\end{align}
where $\gamma$ discounts rewards later in the trajectory. There exist two main strategies to optimise this objective; either directly via the policy gradient \cite{williams1992simple, schulman2017trust}, or indirectly via Q-learning \cite{watkins1989learning, mnih2015human}. We focus on the latter as they tend to be more sample efficient \cite{haarnoja2019soft_, fujimoto2018addressing} given their ability to reuse prior experience and multiple sources of exploration data, a key component of our approach. We describe the specific deep RL algorithm used throughout this work in more detail below.

\subsubsection{Soft-Actor Critic}

SAC is an off-policy, actor-critic algorithm that has achieved state-of-the-art results in recent years for continuous control tasks \cite{haarnoja2019soft_}. It is based on the \textit{maximum entropy} RL framework that optimises a stochastic policy to maximise a trade-off between the expected return and policy entropy, $\mathcal{H}$:
\begin{equation}
\label{entropy_reg}
J(\theta)=\mathbb{E}_{\pi_\theta}\left[\sum_{t=1}^{T} \gamma^{t} r\left(s_{t}, a_{t}\right)+\alpha \mathcal{H}\left(\pi\left(a_{t} \mid s_{t}\right)\right)\right].
\end{equation}

\noindent This is closely related to the exploration-exploitation trade-off \cite{sutton1998introduction}, encouraging exploration in previously unvisited states, where actions are chosen by sampling from a Gaussian distribution. While effective in exhaustively searching the solution space, the associated risk and sample inefficiency are unsuitable for robotics. In this work, we focus on leveraging priors to better inform this action selection process. SAC learns three functions: two Q-functions $Q_{\phi_{1}}$ and $Q_{\phi_{2}}$ that play the role of the critics, and the policy $\pi_{\theta}$, referred to as the actor. The Q-functions are learnt individually by minimising the mean squared bootstrapped estimate (MSBE):

\begin{align}
    \underset{\left(s_{t}, a_t, r, s_{t+1}\right) \sim \mathcal{D}}{\mathbb{E}}\left[\left(Q_{\phi_{i}}(s_{t}, a_t)-y\left(r, s_{t+1}\right)\right)^{2}\right],
\end{align}

\noindent where $\mathcal{D}$ represents the replay buffer and the target, $y\left(r, s_{t+1}\right)$ is augmented with the entropy regularisation term and is given by:
\begin{gather}
\nonumber
\small{y\left(r, s_{t+1}\right)=r+ \gamma\biggl(\min _{j=1,2} Q_{\phi_{\text {targ}, j}}\left(s_{t+1}, \tilde{a}_{t+1}\right) - }\\
\small{\alpha \log \pi_{\theta}\left(\tilde{a}_{t+1} \mid s_{t+1}\right)\biggr),}
\end{gather}

\noindent where $\tilde{a}_{t+1} \sim \pi_\theta(\tilde{a}_{t+1} | s_{t+1})$ is an action sampled from the current policy $\pi_{\theta}$ and the weighting term $\alpha$ explicitly controls the trade-off between reward and entropy maximisation. The policy is a re-parameterised Gaussian with a squashing function to ensure actions are in the range [-1,1]: 
\begin{equation}
a_{\theta_t}\left(s_{t}, \xi_{t}\right)=\tanh \left(\mu_{\theta}\left(s_{t}\right)+\sigma_{\theta}\left(s_{t}\right) \cdot \xi_{t}\right),
\end{equation}

\noindent where $\xi_{t} \sim \mathcal{N}(0, I)$ represents independent sampled noise and $\mu_\theta(s_t)$ and $\sigma_\theta(s_t)$ are the mean and standard deviation of the policy outputs respectively. Finally, the policy parameters are updated by maximising the bootstrapped entropy regularised future return.
\begin{equation}
\small{
\max _{\theta} \underset{\xi \sim \mathcal{N}}{\mathbb{E}}\left[\min _{j=1,2} Q_{\phi_{j}}\left(s_t, {a}_{\theta_t}(s_t, \xi_t)\right)-\alpha \log \pi_{\theta}\left({a}_{\theta_t}(s_t, \xi_t) \mid s_t\right)\right]
}
\end{equation}

\subsection{Control Priors}
\label{sec:control_priors}
The classical robotics community have developed a plethora of algorithms and controllers that are capable of solving various tasks or part of more complex tasks using explicit analytic derivations \cite{siciliano2016springer}. In the context of our work, we refer to these as \textit{control priors}. In their most general form we can view these control priors as deterministic mapping functions from state to action:
\begin{equation}
    a_t = \psi(s_t).
\end{equation}

\begin{figure}[t]
  \centering
  \includegraphics[width=0.5\textwidth]{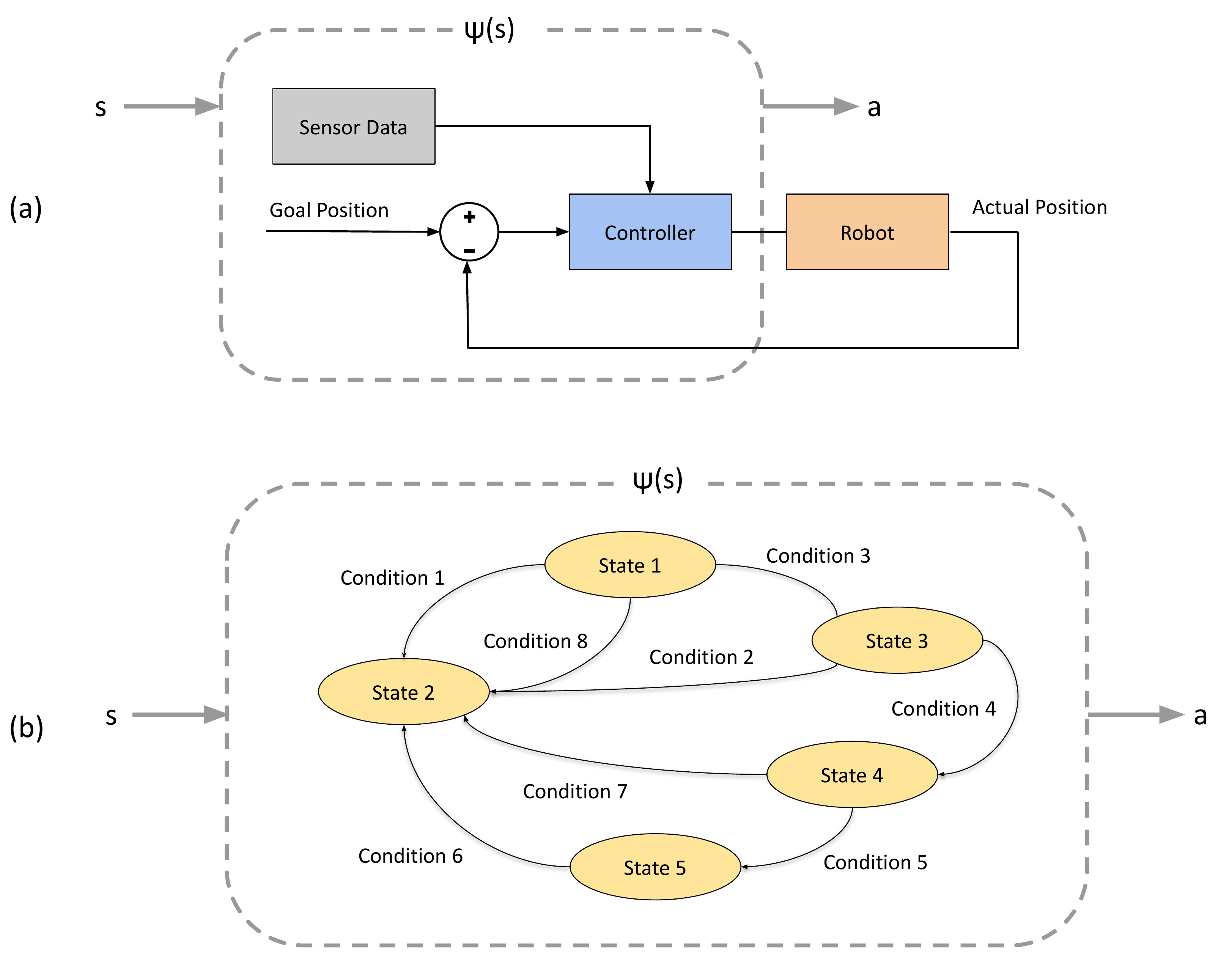}
  \caption{Examples of control priors in robotics, include (a) traditional feedback controllers and (b) state machines. Note that despite the difference in the structure of these priors, they both can be treated as mapping functions, $a=\psi(s)$ from state to action in their most general form.}
  \label{controllers}
\end{figure}

These methods make the assumption that a significant amount is known about the system dynamics, such as the differential equations governing state transitions. Given these explicit models, an error signal can be generated and traditional feedback approaches (e.g. optimal \cite{kirk2004optimal}, model predictive \cite{camacho2013model} and robust control \cite{zhou1998essentials}) can be used to control the input of the system in order to attain the desired behaviours. For more complex and long-horizon robotic tasks, these low-level control systems can be integrated within a larger decision-making framework such as a state machine \cite{Balogh2018UsingFS} or behaviour tree \cite{Colledanchise2017BehaviorTI}. Given their known explicit derivation and deterministic nature, we make the assumption that these controllers are risk-averse and ensure the safety of the robot. This is in contrast to the black-box nature of deep RL policies. Figure \ref{controllers} shows examples of control priors and how we can treat them as mapping functions from states to actions.

These control priors, however, tend to require significant amounts of hand engineering, linear approximations and domain knowledge instilled to ensure that they are functional and can solve the task. As the complexity  of the tasks increases and the operational environment for these controllers becomes unstructured as seen in the real world, such hand-engineered solutions tend to be suboptimal. Explicitly engineering these controllers to meet the level of dexterity required becomes non-trivial and impractical amounts of modelling and calibration may be required.

\section{Related Work}
\label{sec:related_work}
In this section, we review some of the key RLBP methods introduced in the literature. We focus on the specific set of approaches that leverage behavioural priors in the form of control priors within the RL framework and discuss their current limitations when considering their application to robotics.

\subsection{Residual Reinforcement Learning}

The residual reinforcement learning framework \cite{johannink2018residual, silver2018residual, srouji18a} focuses on learning a corrective residual policy for a control prior. The executed action $a_{t}$ is generated by summing the outputs from a control prior and a learned policy, i.e., $a_{t} = \psi(s_{t}) + \pi_{\theta}(s_{t})$. The residual policy, $\pi_\theta$, can be learned using any off-policy RL algorithm. 

\cite{johannink2018residual, silver2018residual} utilise this approach to learn complex manipulation tasks that involve contacts and friction that are typically hard to model analytically. They leverage conventional feedback controllers as control priors that can solve part of the task, and utilise the residual to modify its behaviour in order to learn the unmodelled contact dynamics. Although showing promise towards accelerating the learning of manipulation tasks, they only demonstrate its ability to make small changes to the underlying controller and the implications of utilising highly suboptimal controllers have not yet been explored.

\cite{iscen2018policies} extend this architecture to not only learn a residual but additionally the parameters to modify the classical controller itself. They argue that this allows for a more expressive and flexible architecture, not attainable using solely a residual and shows its applicability to dexterously control a quadrupedal robot. However, allowing the policy direct access to modify the control prior, diminishes any safety guarantees the prior can provide which is a limiting factor when considering the safety of the robot. \cite{srouji18a} attempt to maintain the guarantees of the control prior by restricting the policy to solely learning a linear policy, limiting its ability to make significant modifications to the underlying behaviours of the prior. This strict decoupling additionally allows for a better interpretation of the two controllers, an important factor when assessing the stability and robustness of the overall system. While an interesting perspective, this greatly limits the potential optimality of the final controller that the RL agent can achieve.

\subsection{KL-Regularised Reinforcement Learning}

Recent algorithms directly optimise a regularised objective that trades off reward maximisation with the minimisation of the policy divergence from a behavioural prior which could be a fixed or learned reference distribution. The most common divergence measure used is the Kullback-Leibler (KL) divergence, which computes the degree of similarity (or dissimilarity) between two distributions. The KL-regularised RL objective is given by:
\begin{align}
\label{kl-div}
    J(\theta)=\mathbb{E}_{\pi_\theta}\left[\sum_{t=1}^{T} \gamma^{t} r\left(s_{t}, a_{t}\right)-\alpha D_{\mathrm{KL}}\left(\pi_\theta, \psi\right)\right].
\end{align}

Typically this reference distribution could be a uniform distribution \cite{haarnoja2019soft_}, learned policy \cite{teh2017distral, pertsch2020accelerating, tirumala2019exploiting_, galashov2019information_, hunt2019composing, hausman2018learning}, or an occupancy measure that characterises the distribution of state-action pairs when executing a policy \cite{pmlr-v80-kang18a, jing2020reinforcement}. This approach is typically used for transfer and multi-task learning and has been shown to be effective in distilling prior knowledge into the policy during optimisation.

In the case of a uniform reference distribution, \cite{pertsch2020accelerating} show that we recover the maximum entropy objective given in Equation \ref{entropy_reg}. This is the least informative distribution when considering the incorporation of useful structure and priors to aid the learning process. A more informed distribution constitutes common behaviours \cite{teh2017distral, galashov2019information_} that agents can share across multiple tasks which allows them to significantly accelerate learning. Such behavioural priors are typically learned in parallel or already trained policies that can solve simpler tasks \cite{pertsch2020accelerating}, forming a continual learning paradigm. This is an important motivation for our work, where it makes sense to build on the vast body of work already solved by the robotics community, as opposed to learning policies from scratch.

The KL-constrained setting can be seen as a hard constraint that can severely restrict the policies from attaining optimal behaviours given the potential sub-optimality of the behavioural priors used. \cite{pertsch2020accelerating} address this by automatically learning the weighting parameter, and show the applicability of the regularised objective to accelerate learning of complex tasks within a hierarchical framework. Recent work by \cite{pmlr-v80-kang18a} and \cite{jing2020reinforcement} softens the constraint by gradually annealing the divergence tolerance. This allows the policy to deviate further away from the prior as training progresses in order to learn potentially better behaviours, an important factor when the prior is highly suboptimal. The choice of the annealing rate is however non-trivial and needs to be carefully selected.

While overall a promising approach to accelerate learning, the KL regularised objective provides limited safety guarantees to the agent during training. This is particularly important in the robotics case, where we additionally seek to ensure the safety of the robot during exploration. A more promising strategy would be to directly utilise the behavioural prior to influence the actions taken by the agent during exploration. We discuss these approaches below.

\subsection{Exploration Bias using Behavioural Priors}

These methods can be interpreted as biasing the policy search towards the behavioural prior's actions during exploration. This is done by utilising the prior as a source of exploration data, directly moderating the agent's behaviours in the given environment.

The simplest approach to this method is policy reuse/intertwining proposed by \cite{Fernndez2006ProbabilisticPR, jeong2020learning}, who utilise an epsilon-greedy-like approach to balance exploration and the usage of a behavioural prior. While providing an attractive solution, this method is not suitable when considering the robotics scenario. Such an approach alternates between the policy, prior and random exploration throughout training without any concern for the safety of these actions. Taking completely random actions at the start of training can result in unsafe behaviours that can be detrimental to both the robot and its surroundings.

An alternative approach is to restrict exploration to be governed by the behavioural prior only until the policy is capable of safely exploring on its own accord. \cite{rana_mcf} proposed a multiplicative fusion strategy for stochastic policies and behavioural priors. This approach utilises a gating function to initially allow exploratory actions only to be sampled from the behavioural prior and gradually transition control towards the policy in order to exploit its own actions. Tuning the gating function can be tedious, and requires the correct balance, while the hard switch towards the policy has limited safety guarantees since the behavioural prior has no impact after this point. This can result in unsafe behaviours as the policy continues to explore novel regions of the solution space on its own. In this work we build on this formulation, exploring alternatives to the hard-gating function in order to address this limitation and allow for continual safe exploration.

A better-informed strategy is to govern this transition based on the policy's confidence. \cite{cheng2019control} utilise the temporal difference error (TD-error) to estimate how confident the policy is to act in a given state. This, in theory, should allow the control prior to dominate exploration during the early stages of training and as the TD-error reduces, the policy can gradually take over. The TD-error however tends to be noisy in practice and can yield instabilities during training. In this work, we explore the use of epistemic uncertainty estimation techniques for RL as a more stable alternative to the TD-error estimate. \cite{xie2018learning} propose a separately trained critic network to govern this arbitration. While this allows for expressive compositions of the two systems, the requirement to learn a separate critic increases the sample complexity of the overall approach, with limited ability to deal with out-of-distributions states. These are core limitations we address in this work, where both accelerating training and the safety of the agent are important factors to consider for robotic systems.

Given the distribution mismatch between the current policy and the behavioural prior, an important factor to consider with these approaches is the exploration-exploitation trade-off between the two sources of information. Excessive reliance on the behavioural prior for experience collection, without adequately exploiting the policy's behaviours can result in unstable learning due to \textit{extrapolation error} \cite{kumar2019stabilizing, fujimoto2018addressing, van2016deep}. Extrapolation error tends to occur as a result of the target Q value estimate, $r+\gamma Q\left(s_{t+1}, a_{t+1}\right)$ that is used to update the value network, where $a_{t+1}$ is selected by the current policy $\pi(s_{t+1})$. However if all the exploration data is collected using the behavioural prior and the state-action pair $(s_{t+1}, a_{t+1})$ is not contained in the training set, the value $Q(s_{t+1}, a_{t+1})$ can be unpredictable. This is due to extrapolation from other state-action pairs as a result of function approximation using neural networks. We explore a novel uncertainty-based strategy to balance exploration with the two systems in order to avoid this instability.\\

Given the range of RLPB approaches discussed above and their respective limitations towards applicability in the robotics setting, we formulate the problem in Section \ref{sec:pf} before presenting our approach to address these limitations in Section \ref{sec:bcf}.

\section{Problem Formulation}
\label{sec:pf}
Decades of research and development have resulted in algorithmic solutions for a number of problems in robotics, distilling analytical approaches, domain knowledge, and human intuition \cite{siciliano2016springer}. Instead of ignoring this vast resource and learning robotic tasks from scratch, we present a control strategy that seamlessly combines a learned policy $\pi$ with an existing hand-crafted algorithm $\psi$, during both training \emph{and} deployment. 
We refer to $\psi$ as a \emph{control prior}. We assume that $\psi$ is suboptimal, but has a reasonable degree of task competence and is risk-averse. We define risk aversion as the tendency of these controllers to avoid unsafe behaviours that could be detrimental to both the safety of the robot and its surroundings.

We consider learning a policy $\pi(a|s)$ for a robotics task. Using the formalism of an MDP, we leverage off-policy model-free RL to learn this policy. However, in contrast to most existing RL approaches, we exploit the knowledge and structure encoded in an existing prior $\psi(a|s)$ during both training \emph{and} deployment for accelerated learning and safer behaviours.

Our goals are to (1) use the control prior $\psi$ to guide exploration during the early phases of the learning process, thereby accelerating training and improving sample efficiency; (2) naturally let the learned policy $\pi$ dominate control as it gains more knowledge, ensuring it can \emph{improve} beyond the performance of the existing solution $\psi$ and (3) monitor the \emph{uncertainty} of $\pi$ during deployment and smoothly transfer control to $\psi$ as a safe-but-suboptimal fall-back in situations where the learned policy cannot be trusted. In Section~\ref{sec:bcf} we explain how a Bayesian combination of $\pi$ and $\psi$ can achieve the above goals. 



\section{Bayesian Controller Fusion}
\label{sec:bcf}

We introduce Bayesian Controller Fusion (BCF), a hybrid control strategy that composes stochastic action outputs from two separate control mechanisms: an RL policy $\pi(a|s)$, and a control prior $\psi(a|s)$. These outputs are formulated as distributions over actions, where each distribution captures the uncertainty over the selected action in any given state. The Bayesian composition of these two outputs forms our hybrid policy $\phi(a|s)$. We show that governing the agent's behaviours in the environment using $\phi(a|s)$, significantly accelerates learning towards an optimal policy while allowing for safer exploration and operation in unknown environments. We describe the intuition behind BCF in more detail below.

\paragraph*{\textbf{Accelerated Learning:}}

\begin{figure}[t]
  \centering
  \includegraphics[width=0.49\textwidth]{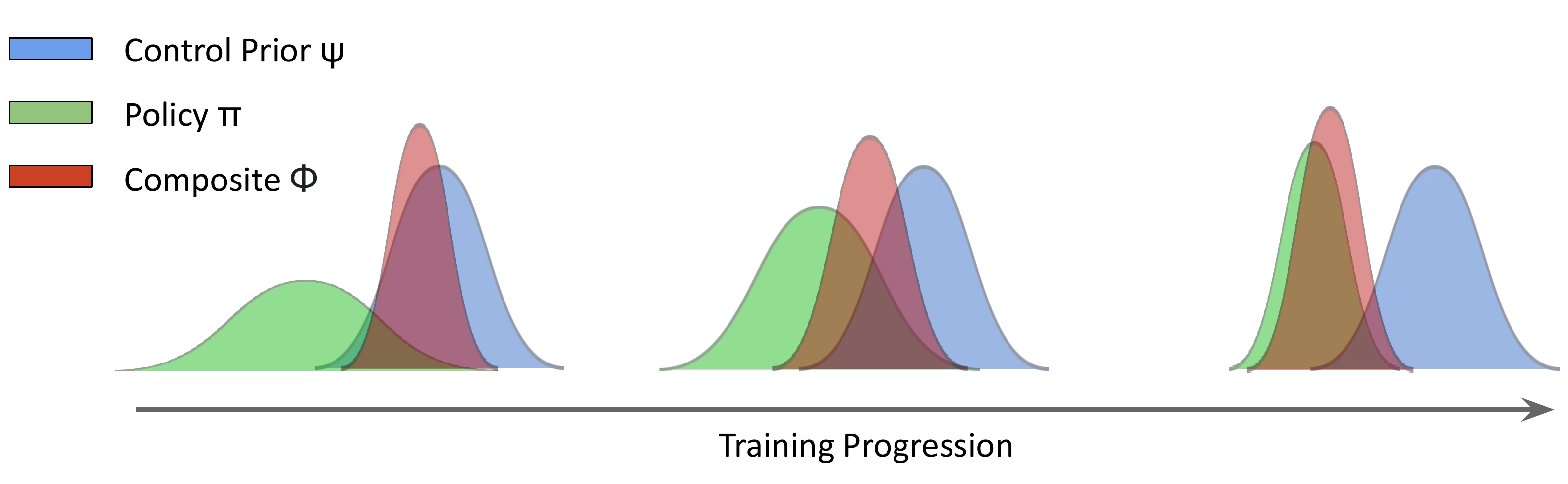}
  \caption{Evolution of the composite BCF sampling distribution over the course of training. During the early stages, note the strong bias towards the control prior which provides the exploration guidance.}
  \label{gaussians}
\end{figure}

As the action distribution generated by each system captures the uncertainty over the selected action in a given state, our Bayesian fusion strategy allows the composite distribution $\phi(a|s)$ to naturally bias towards the system exhibiting the least amount of uncertainty. Figure \ref{gaussians} shows the evolution of our exploration strategy throughout training. During the early stages of training, the policy $\pi(a|s)$ exhibits high uncertainty across all states, biasing the composite distribution towards the control prior $\psi(a|s)$. As opposed to random exploration, sampling actions from this distribution allows the control prior to strongly influence exploration at this stage. This quickly biases the agent towards the reward-yielding trajectories, while exploring the surrounding state-action space for potential improvements beyond the sub-optimality of the control prior. As the policy becomes more certain about its environment it gradually dominates control, allowing the agent to exploit its learned behaviours and stabilise training.

\paragraph*{\textbf{Safe Exploration and Deployment:}} 

\begin{figure}[t]
  \centering
  \includegraphics[width=0.49\textwidth]{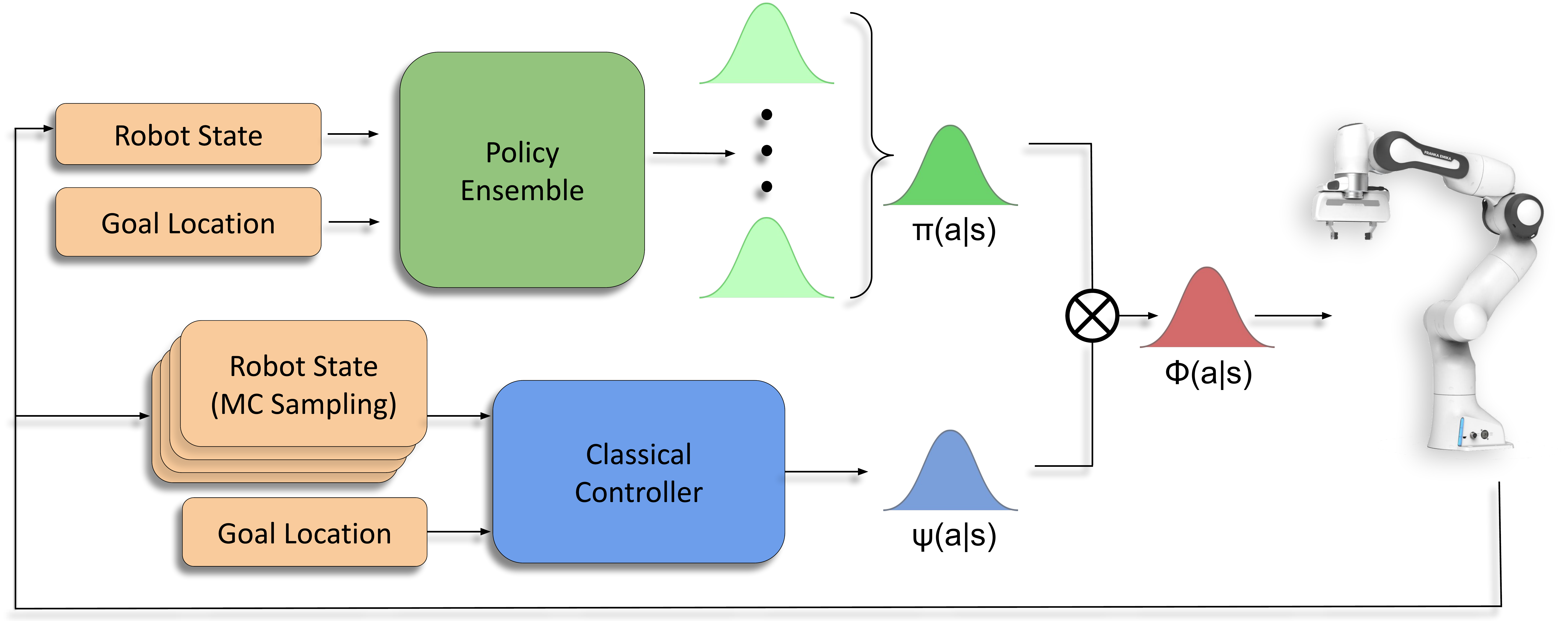}
  \caption{BCF hybrid control strategy for safe deployment on real robotic systems. We derive uncertainty-aware action outputs for each controller and compose these outputs to better inform the action selection process.}
  \label{front}
\end{figure}

Figure \ref{front} illustrates our hybrid control strategy that composes the outputs from the learned policy $\pi(a|s)$ and control prior $\psi(a|s)$. The uncertainty-aware compositional policy $\phi(a|s)$ allows for the safe exploration and deployment of RL agents. In states of high uncertainty, the compositional distribution naturally biases towards the reliable, risk-averse and potentially suboptimal behaviours suggested by the control prior. In states of lower uncertainty, it biases towards the policy, allowing the agent to exploit the optimal behaviours discovered by it. This is reminiscent of the arbitration mechanism suggested by \cite{lee2014neural} for behavioural control in the human brain, where the most reliable controller influences control in a given situation. This dual-control perspective provides a reliable strategy for bringing RL to real-world robotics, where generalisation to all states is near impossible and the presence of risk-averse control priors serves as reliable fallback systems.

\subsection{Method}
\label{method}

Given a policy, $\pi$ and control prior, $\psi$, we can obtain two independent estimates of an executable action, $a$ conditioned on both a given state $s$ and the task $T$.
In a Bayesian context, we can utilise the normalised product to fuse these estimates under the assumption of a uniform prior, $p(a)$:
\begin{align}
    \label{eq_1}
    p\left(a \mid \mathbf{\theta}_{\pi}, \mathbf{\theta}_{\psi}\right)=\frac{p\left(\mathbf{\theta}_{\pi}, \mathbf{\theta}_{\psi} \mid a\right) p(a)}{p\left(\mathbf{\theta}_{\pi}, \mathbf{\theta}_{\psi}\right)},
\end{align}

\noindent where we assume Gaussian distributional outputs from each system and represent $\pi(a|s,T) \sim p\left(a\mid s, T, \theta_{\pi}\right)$ and $\psi(a|s,T) \sim p\left(a\mid s, T, \theta_\psi\right)$, where, $\theta_{\pi} = \{[\mu_{\pi_1},...,\mu_{\pi_n}]^\intercal, [\sigma_{\pi_1},...,\sigma_{\pi_n}]^\intercal\}$ and $\theta_{\psi} = \{[\mu_{\psi_1},...,\mu_{\psi_n}]^\intercal, [\sigma_{\psi_1},...,\sigma_{\psi_n}]^\intercal\}$ denote the distribution parameters for the policy and control prior outputs respectively, $\mu$ and $\sigma$ denote the mean and standard deviation and $n$ is the dimensionality of the action space. We drop the state $s$ and task $T$ to simplify the notation.\\

\noindent Assuming statistical independence of $p\left(\mathbf{\theta}_{\pi} \mid a\right)$  and $p\left(\mathbf{\theta}_{\psi} \mid a\right)$, we can expand our likelihood estimate, $p\left(\mathbf{\theta}_{\pi}, \mathbf{\theta}_{\psi} \mid a\right)$, as follows:
\begin{align}
\label{eq_2}
    p\left(\mathbf{\theta}_{\pi}, \mathbf{\theta}_{\psi} \mid a\right)&=p\left(\mathbf{\theta}_{\pi} \mid a\right) p\left(\mathbf{\theta}_{\psi} \mid a\right)\nonumber \\
    &=\frac{p\left(a \mid \mathbf{\theta}_{\pi}\right) p\left(\mathbf{\theta}_{\pi}\right)}{p(a)} \frac{p\left(a \mid \mathbf{\theta}_{\psi}\right) p\left(\mathbf{\theta}_{\psi}\right)}{p(a)}.
\end{align}

\noindent Substituting this result back into (\ref{eq_1}), we can simplify the fusion as a normalised product of the respective action distributions from each control mechanism:

\begin{align}
  p\left(a \mid \mathbf{\theta}_{\pi}, \mathbf{\theta}_{\psi}\right)=\eta \underbrace{p\left(a \mid \mathbf{\theta}_{\pi}\right)}_{\substack{\text{Policy}}} \underbrace{p\left(a \mid \mathbf{\theta}_{\psi}\right)}_{\substack{\text{Control} \\ \text{Prior}}},
\end{align}
where,
\begin{align}
  \eta=\frac{p\left(\mathbf{\theta}_{\pi}\right) p\left(\mathbf{\theta}_{\psi}\right)}{p\left(\mathbf{\theta}_{\pi}, \mathbf{\theta}_{\psi}\right) p(a).} 
\end{align}



\noindent The composite distribution $p\left(a \mid \mathbf{\theta}_{\pi}, \mathbf{\theta}_{\psi}\right)$ forms our hybrid policy output\ $\phi(a|s)$. As we approximate the distributional output from each system to be univariate Gaussian for each action, the composite distribution $\phi(a|s)$ will also be univariate Gaussian $\phi(a|s) \sim \mathcal{N}(\mu_\phi, \sigma^{2}_\phi)$. As a result, we can compute the corresponding mean $\mu_{\phi}$ and variance $\sigma^{2}_{\phi}$  for the distribution:
\begin{equation}
    \label{h_mu}
   \mu_\phi = \frac{\mu_{\pi}\sigma_{\psi}^{2} + \mu_{\psi}\sigma_{\pi}^{2}}{\sigma_{\psi}^{2} + \sigma_{\pi}^{2}} ,
\end{equation}

\begin{equation}
    \label{h_sig}
    \sigma_\phi^{2} = \frac{\sigma^{2}_{\pi}\sigma_{\psi}^{2}}{\sigma_{\psi}^{2} + \sigma_{\pi}^{2}} , 
\end{equation}

\noindent where this expansion implicitly handles the normalisation constant $\eta$. 

\subsection{Components}
\label{sec:comp}
In order to leverage our proposed approach in practice, we describe the derivation of the distributional action outputs for each system below and provide the complete BCF algorithm for combining these systems in Algorithm \ref{algorithm1}.

\subsubsection{Uncertainty-Aware Policy}
 We leverage stochastic RL algorithms that output each action as an independent Gaussian $\pi'(a|s)\sim \mathcal{N}(\mu_{\pi'}, \sigma_{\pi'}^2)$ where $\mu_{\pi'}$ denotes the mean and $\sigma^{2}_{\pi'}$ denotes the corresponding variance. This distribution is optimised to reflect the action which would both maximise the returns from a given state, as well as the entropy  \cite{haarnoja2019soft_}. Such exploration distributions tend to be risk-seeking and do not capture the uncertainty over the actions selected by the agent. The latter is a key component required for our BCF formulation. To attain an uncertainty-aware distribution, we leverage \textit{epistemic} uncertainty estimation techniques suggested in the computer vision literature based on ensemble learning \cite{lakshminarayanan2017simple}.
 
 We train an ensemble of $M$ agents to form a uniformly weighted Gaussian mixture model, and combine these predictions into a single univariate Gaussian whose mean and variance are respectively the mean, $\mu_{\pi}(s)$  and variance, $\sigma^{2}_{\pi}(s)$ of the mixture, $p(a \mid s, \theta_{\pi})=M^{-1} \sum_{m=1}^{M} p\left(a \mid s, \theta_{\pi'_m}\right)$. The mean and variance of the mixture $M^{-1} \sum \mathcal{N}\left(\mu_{\pi'_m}(s), \sigma_{\pi'_m}^{2}(s)\right)$ are given by:
 \begin{equation}
 \label{fuse_mean}
\mu_{\pi}(s)=M^{-1} \sum_{m} \mu_{\pi'_{m}}(s),
\end{equation}

\begin{equation}
\label{fuse_sigma}
\sigma_{\pi}^{2}(s)=M^{-1} \sum_{m}\left(\sigma_{\pi'_{m}}^{2}(s)+\mu_{\pi'_{m}}^{2}(s)\right)-\mu_{\pi}^{2}(s).
\end{equation}
 
\noindent The empirical variance, $\sigma_{\pi}^{2}(s)$, of the resulting output distribution, $p(a \mid s, \theta_{\pi})$ approximates a measure of the policy's epistemic uncertainty over actions in a given state. This allows for a broader distribution when presented with unknown states and a tighter distribution in familiar states. This plays an important role in our BCF formulation as described previously. We note here that we do not employ the additional adversarial training step suggested by \cite{lakshminarayanan2017simple} when generating the ensembles as we utilise low-dimensional state representations as opposed to images as described in Section \ref{sec:tasks}.

\subsubsection{Control Prior}

 In order to incorporate the inherently deterministic control priors developed by the robotics community within our stochastic RL framework, we require a distributional action output that captures its uncertainty over actions in a given state.  We empirically derive this action distribution by propagating noise (provided by the known sensor model variance, $\sigma^{2}_{\text{model}}$) from the sensor measurements through to the action outputs using Monte Carlo (MC) sampling. We make the assumption that the noise induced by this sensor is small and does not impact the full observability of state information provided to the RL agent. By computing the mean, $\mu_{\psi}$ and variance, $\sigma^{2}_{\psi}$ of the outputs, the distributional action output, $\mathcal{N}(\mu_{\psi}, \sigma_{\psi}^{2})$ for a given state, $s$ is given by:
 
\begin{equation}
    \label{cp_mu}
    \mu_{\psi}(s)=N^{-1} \sum_{n} a_{\psi}(s_{MC_{n}}), \hspace{20pt} s_{MC} \sim \mathcal{N}(s, \sigma_{\text{model}}^{2}),
\end{equation}

\begin{equation}
    \sigma_{\psi}^2(s) = N^{-1} \sum_{n} (a_{\psi}(s_{MC_{n}}) - \mu_{\psi}(s))^2,
\end{equation}
 
\noindent where $a_{\psi}(\cdot)$ denotes a deterministic action output from the control prior for a given state and $N$ is the number of sampled states. Given that some control priors are inherently robust to noise or the sensors used may exhibit minimal noise, we additionally set a minimum possible standard deviation for the distribution. This prevents the control prior distribution from collapsing to a deterministic value which would render the policy useless within the BCF formulation. The resulting variance for the control prior distribution is defined as:
 
 \begin{align}
 \label{cp_distr}
    \sigma_{\psi}^2(s) =  \max\left( N^{-1} \sum_{n} (a_{\psi}(s_{MC_{n}}) - \mu_{\psi}(s))^2, \sigma_{d}^{2}(s)\right).
\end{align}
 
\begin{figure}[t]
  \centering
  \includegraphics[width=0.49\textwidth]{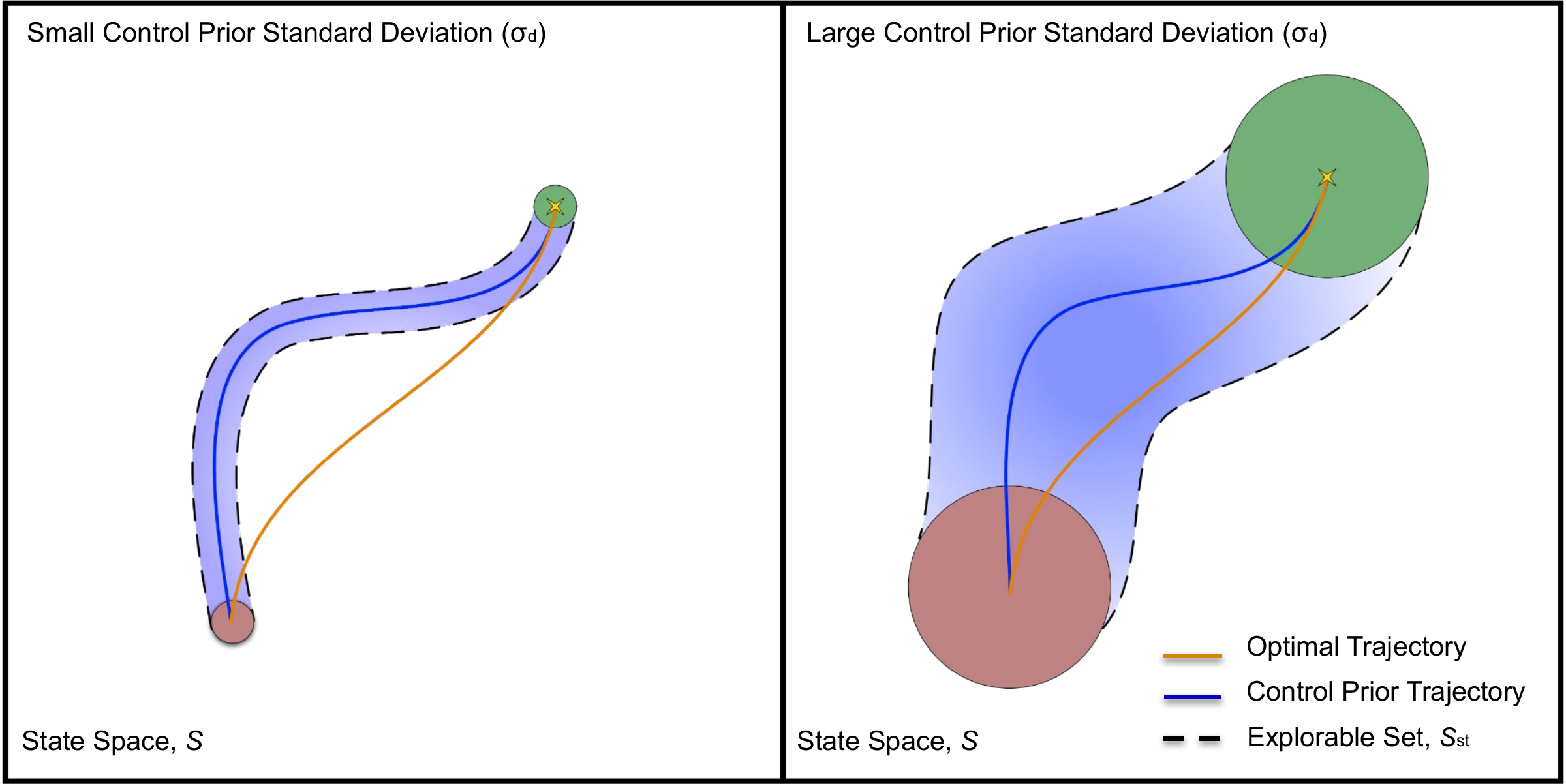}
  \caption{Illustration of optimal trajectory vs. control prior trajectory with the explorable set $\mathcal{S}_{st}$ extracted. The chosen standard deviation, $\sigma_d$ should reflect the optimality of the controller such that the optimal behaviours can be captured within the distribution.}
  \label{control_reg}
\end{figure}

 \noindent The choice of $\sigma_d$ is left as a hyper-parameter for the user to set based on the specific controller used and its optimality towards solving the task. Figure \ref{control_reg} provides some intuition into the choice of $\sigma_d$ and its relation to exploration. The explorable region of the state space is denoted by the set $\mathcal{S}_{st}$, which grows as $\sigma_d$ is increased and vice versa. Sampling actions from this trust region depicts the guided exploration provided by the control prior. The ability to bias the policy towards the optimal policy depends on this explorable set. If the optimal trajectory is within the explorable region, then we can learn the corresponding optimal policy - otherwise, the policy will remain suboptimal. We conduct an ablation study to further explore the impact that this hyper-parameter has during training in Section \ref{sec:ablation_variance}.\\

Given the formulation for the distributional outputs from each system, we present the complete BCF algorithm for governing action selection both during training and deployment in Algorithm \ref{algorithm1}. 

\section{Experimental Setup}
\label{sec:experiments_setup}
In this section, we describe the two different robotics tasks that we evaluate BCF on and detail the analytic derivation of the control priors we use for these tasks and their respective limitations. We additionally provide specific implementation details pertaining to our training method. We note here that the control priors used are just one particular example of existing hand-crafted solutions to the tasks, and alternative approaches could also be used.

\subsection{Tasks}
\label{sec:tasks}
\begin{figure}[t]
  \centering
  \includegraphics[width=0.49\textwidth]{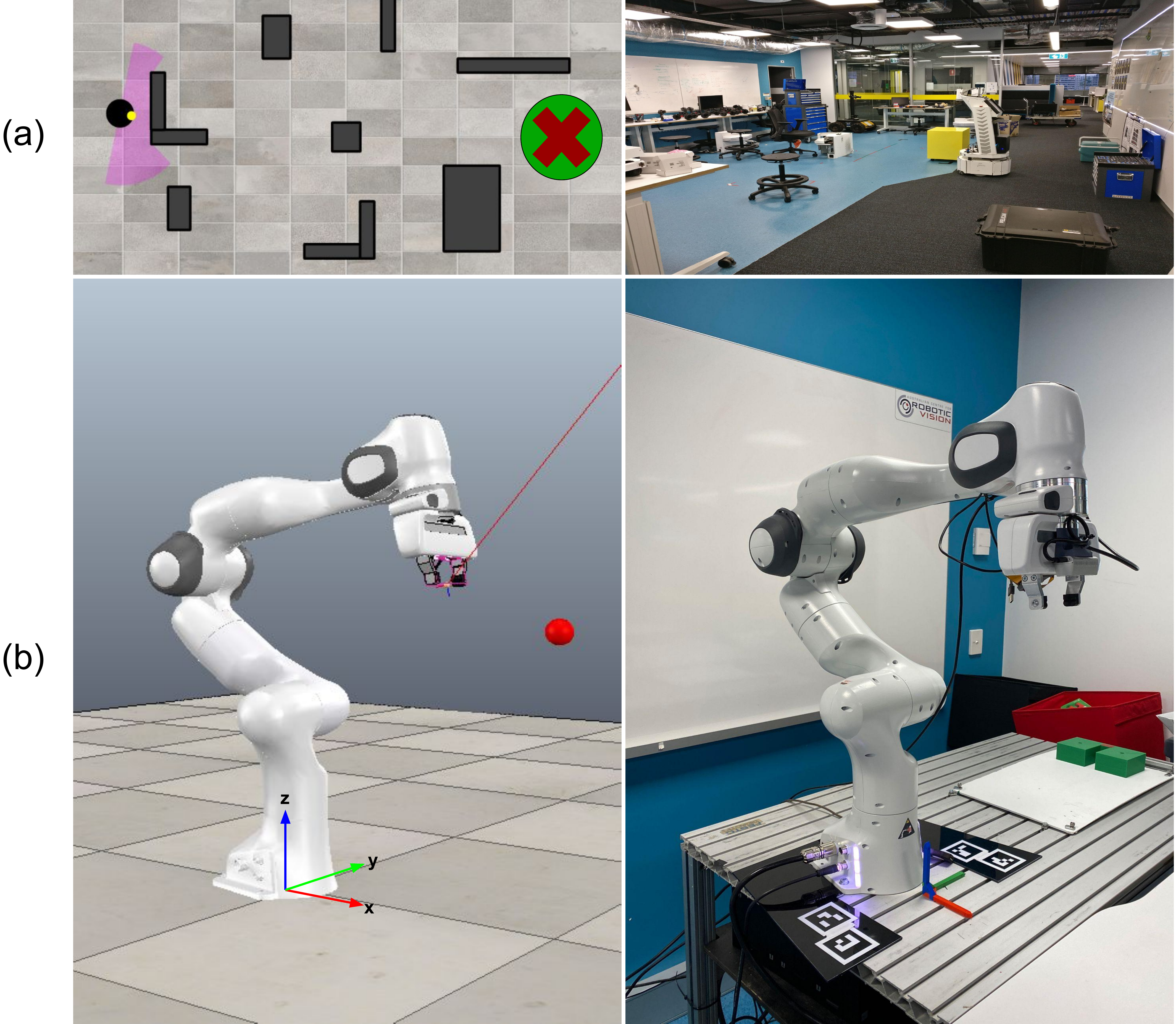}
  \caption{Simulation training environments and real-world deployment environments for (a) PointGoal Navigation and (b) Maximum Manipulability Reacher tasks. Note the stark discrepancy in obstacle profiles for the navigation task between the simulation environment and real-world environments.}
  \label{robots}
\end{figure}

We study two different task domains (shown in Figure \ref{robots}) to evaluate the performance of BCF as a viable strategy towards exploiting the strengths of RL and classical control, both during training and real-world deployment. To provide a direct comparison with previous work \cite{rana_mcf}, we conduct experiments in the \textit{PointGoal Navigation} task which was first presented by \cite{anderson2018evaluation} as well as a complex reaching task that requires manipulability maximisation across the trajectory. For both tasks, we assume the sparse reward, long horizon setting and the presence of existing control priors that can provide some structure towards solving the task. We describe each task in more detail below.

\begin{algorithm}[t]
\SetAlgoLined
\textbf{Given:} Ensemble of \textit{M} policies ($[\pi'_{1}, \pi'_{2} ... \pi'_{M}]$), control prior ($\psi$) and default control prior variance ($\sigma_d^{2}$) \\
\KwIn{State $\textit{s}_t$}
\KwOut{Action $\textit{a}_t$}
  Approximate the policy ensemble predictions as a unimodal Gaussian $\pi(\cdot|s_{t}) \sim \mathcal{N}(\mu_{\pi}, \sigma_{\pi}^{2})$ described in Equations (\ref{fuse_mean}) and (\ref{fuse_sigma})\\
  
  Compute the control prior action distribution $\psi(\cdot|s_{t}) \sim \mathcal{N}\left(\mu_{\psi}, \sigma_{\psi}^{2}\right)$ as given in Equations (\ref{cp_distr}) and (\ref{cp_mu}) \\
  
  Compute the composite distribution $\phi(\cdot|s_{t}) \sim \mathcal{N}(\mu_\phi, \sigma^{2}_\phi)$\\
  
  \nonl \hspace{0.1cm} $\phi(\cdot|s_{t}) = \eta(\pi(\cdot|s_{t})\cdot\psi(\cdot|s_{t}))$ as given in Equations (\ref{h_mu}) and (\ref{h_sig})\\
  
  Select action $\textit{a}_t$ from the distribution $\phi(\cdot|s_{t})$\\
  
 \Return{$\textit{a}_t$}
 \caption{Bayesian Controller Fusion}
\label{algorithm1}
\end{algorithm}

\paragraph*{\textbf{PointGoal Navigation:}} 
The objective of this task is to navigate a robot from a start location to a goal location in the shortest time possible while avoiding obstacles along the way. We utilise the training environment provided by \cite{rana2019residual}, which consists of five arenas with different configurations of obstacles. The goal and start location of the robot are randomised at the start of every episode, each placed on the extreme opposite ends of the arena (see Figure~\ref{robots} (a)). This sets the long-horizon nature of the task. As we focus on the sparse reward setting, we define $r(s_t,a_t,s_{t+1}) = 1$ if $d_{\text{target}} < d_{\text{threshold}} $ and $r(s_t,a_t,s_{t+1}) = 0$ otherwise, where $d_{\text{target}}$ is the distance between the agent and the goal and $d_{\text{threshold}}$ is a set threshold. The action $a_t$ consists of two continuous values: linear velocity $\nu_{\text{nav}}\in[-1,1]$ and angular velocity $\omega_{\text{nav}}\in[-1,1]$. We assume that the robot can localise itself within a global map in order to determine its relative position to a goal location. The 180$^{\circ}$ laser scan range data is divided into 15 bins and concatenated to the robot's angle. The overall state $s_{t}$ of the environment is comprised of:
\begin{itemize}
    \item The binned laser scan data $l_{\text{bin}} \in \mathbb{R}^{15}$,
    \item The polar pose error between the robot's pose and the goal location $e_{t} \in \mathbb{R}^2$,
    \item The previous executed linear and angular velocity $a_{t-1} \in \mathbb{R}^2$,
\end{itemize}
for a total of 19 dimensions. The length of each episode is set to a maximum of 500 steps and does not terminate once the goal is achieved.

\paragraph*{\textbf{Reaching with Maximum Manipulability:}}
The objective of this task is to actuate each joint of a manipulator within a closed-loop velocity controller such that the end-effector moves towards and reaches the goal point, while the manipulability of the manipulator is maximised. The manipulability index describes how easily the manipulator can achieve any arbitrary velocity. The ability of the manipulator to achieve an arbitrary end-effector velocity is a function of the manipulator Jacobian. While there are many methods that seek to summarise this, the manipulability index proposed by \cite{manip} is the most used and accepted within the robotics community \cite{manip2}. Utilising Jacobian-based indices in existing controllers have several limitations, requiring greater engineering effort than simple inverse kinematics-based reaching systems, and precise tuning in order to ensure the system is operational \cite{mmc}. We explore the use of RL to learn the desired behaviours by leveraging simple reaching controllers as priors. We utilise the PyRep simulation environment \cite{james2019pyrep}, with the Franka Emika Panda as our manipulator as shown in Figure \ref{robots} (b).  For this task, we generate a random initial joint configuration and random end-effector goal pose. We use a sparse goal reward, $r(s_t,a_t,s_{t+1}) = 1$ if $e_{t} < e_{\text{threshold}} $ and $r(s_t,a_t,s_{t+1}) = m$ otherwise, where $e_{t}$ is the spatial translational error between the end-effector and the goal, $e_{\text{threshold}}$ is a set threshold and
\begin{equation}
    m = \sqrt{ 
            \mbox{det}
            \left(
                J(q) J(q)^\top
            \right) 
    }  \ \in [0,\infty)
\end{equation}
is the manipulability of the robot at the particular joint configuration $q$ where $J(q)$ is the manipulator Jacobian. The action space consists of the manipulator joint velocities $\dot{\it{q}} \in [-1,1]^n$, where the values are continuous, and $n$ is the number of joints within the manipulator. In this work, the manipulator used consists of 7 joints. The state, $s_t$, of the environment is comprised of:
\begin{itemize}
    \item The joint coordinate vector $\it{q} \in \mathbb{R}^7$,
    \item The joint velocity vector $\dot{\it{q}} \in \mathbb{R}^7$,
    \item The translation error between the manipulator's end-effector and the goal $e_t(\it{q}) \in \mathbb{R}^3$,
    \item The end-effector translation vector $e_p(\it{q}) \in \mathbb{R}^3$,
\end{itemize}
for a total of 20 dimensions. Similar to the navigation task, the episode length for this task in fixed at 1000 steps and only terminates at the end of the episode.


\subsection{Control Prior Derivation}
\label{classical_controllers}

For each of the above tasks, we utilise existing classical controllers and algorithms already developed by the robotics community for solving the task at hand. They however are not necessarily optimal, due to limitations in analytical modelling, controller miscalibration and task variations. We note here that each controller was calibrated, and scripted conditions were implemented to ensure they exhibited safe and risk-averse behaviours across the state space. We describe each of these controllers in more detail below.

\paragraph*{\textbf{Artificial Potential Fields:}}
\label{control_prior_APF} Assuming the availability of global localisation and map information for our robot, \textit{PointGoal Navigation} can mostly be solved using classical reactive navigation techniques. These systems rely on the immediate perception of their surrounding environment which allows them to handle dynamic objects and those unaccounted for in the global map. In this work we focus on the Artificial Potential Fields (APF) family of algorithms \cite{warren1989global, koren1991potential} which compute a local attractive potential $P_g \in \mathbb{R}$ that attracts the robot at location $(x_{\text{robot}},y_{\text{robot}})$ towards the goal $(x_{\text{goal}}, y_{\text{goal}})$,  while a repulsive potential $P_o \in \mathbb{R}$ repels it away from obstacles $o$:
\begin{equation}
P_{g}=\frac{1}{2} \zeta_1\sqrt{\left|x_{\text{robot}}-x_{\text {goal }}\right|^{2}+\left|y_{\text{robot}}-y_{\text {goal }}\right|^{2}}
\end{equation}

\begin{equation}
P_o =  \frac{1}{2} \zeta_2\left(\frac{1}{D(o)}\right)^{2},
\end{equation}

\noindent where $\zeta_1 \in \mathbb{R}^+$ and $\zeta_2 \in \mathbb{R}^+$ are gain terms, and $D(o) \in \mathbb{R}$ is a function of the obstacle dimensions and their distance from the robot, $o \in \mathbb{R}^j$, where $j$ is the number of obstacles. $\zeta_2$ can be tuned such that the robot is always a safe distance away from obstacles, ensuring that it does not experience collisions. The resulting potential function $P_r$ is then computed by combining these components:
\begin{align}
    P_r&=P_g + P_o
\end{align}

\noindent By taking the gradient of the resultant potential function, we can determine the best direction to move within the environment that avoids obstacles while heading towards the goal. 
\begin{align}
    F&=-\nabla P_r
\end{align}

\noindent For a detailed derivation, we refer the reader to \cite{koren1991potential}. This direction is used to generate an error signal $e_p$ with which we derive a linear feedback controller $u = K_{\text{nav}}e_p$ to determine the suitable velocities $u$ to control the robot, where $K_{\text{nav}} \in \mathbb{R}^+$ is the proportional gain. We implement a variant of this algorithm that utilises only the direct information provided by a laser scanner with no additional global information about the obstacles in the environment. This is a typical scenario in real-world dynamic environments where surroundings may constantly change and the robot has to quickly respond to these changes.

A key problem faced by most classical solutions to reactive navigation, including APF, is the need for extensive tuning and hand engineering to achieve good performance, and a tendency to deteriorate in performance when overfit to a particular region \cite{koren1991potential,khatib1986real}. This makes them susceptible to oscillations, seizure in local minima, and suboptimal path efficiency.

\paragraph*{\textbf{Resolved Rate Motion Control:}}

This controller allows for direct control of a robotic manipulator's end-effector, without expensive path planning. It serves as a simple prior for a reaching task. For a given goal pose, Resolved Rate Motion Control (RRMC) provides the robot with suitable joint velocity commands to move the end-effector in a straight line towards the goal.

For a given manipulator joint configuration $\it{q}\in \mathbb{R}^n$ where $n$ is the number of joints within the manipulator, we can calculate the forward kinematics through the non-linear surjective mapping 
\begin{equation} \label{m:rrmc0}
    p=f(q), 
\end{equation}
where $p \in \R^m$ is some parameterisation of the end-effector pose, and the mapping function $f(\cdot)$ is a function of the robot's geometric structure. We are using a manipulator with a task space $\mathcal{T} \in \SE{3}$, and therefore $m = 6$. Taking the time derivative of (\ref{m:rrmc0}) gives:
\begin{equation}  \label{m:rrmc1}
    \nu_{\text{man}} = J(q)\dot{q}, 
\end{equation}
where $J(q) = \frac{\partial f(q)}{\partial q} \big{|}_{q=q_i} \in \R^{6 \times n}$ is the manipulator Jacobian for the robot at joint configuration $q$. RRMC exploits the mapping between a Cartesian end-effector velocity $\nu_{\text{man}}$ to the manipulator's joint velocity $\dot{q}$ provided by the differential kinematics in (\ref{m:rrmc1}) \cite{rrmc}. By rearranging (\ref{m:rrmc1}), the required joint velocities to achieve an arbitrary end-effector velocity can be calculated as:
\begin{equation}  \label{m:rrmc2}
    \dot{q} = J(q)^{-1} \ \nu_{\text{man}},
\end{equation}
(\ref{m:rrmc2}) can only be solved when $J(q)$ is square and non-singular. For redundant robots (where $n>6$), $J(q)$ is not square and therefore no unique solution exists for (\ref{m:rrmc2}). Consequently, the most common solution is to use the Moore-Penrose pseudo-inverse in (\ref{m:rrmc2}) as follows:
\begin{equation} \label{m:rrmc3}
    \dot{q} = J(q)^{+} \ \vec{\nu_{\text{man}}},
\end{equation}
where $(\cdot)^+$ denotes the pseudo-inverse operation. The pseudo-inverse will find $\dot{q}$ with the minimum Euclidean norm. RRMC can be wrapped into a closed-loop velocity controller using position-based servoing (PBS). PBS seeks to drive the robot's end-effector in a straight line from its current pose to the desired pose. This control scheme is formulated as
\begin{equation} \label{eq:pbs}
    \vec{\nu}_e = K_{\text{man}} \ \beta\left( (\mat[0]{T}_e)^{-1} \bullet \mat[0]{T}_{e^*} \right)
\end{equation}
where $K_{\text{man}} \in \mathbb{R}^+$ is a gain term, $\mat[0]{T}_e \in \SE{3}$ is the current end-effector pose expressed in the robot's base frame, $\mat[0]{T}_{e^*} \in \SE{3}$ is the desired end-effector pose expressed in the robot's base frame, and $\beta(\cdot): \mathbb{R}^{4\times 4} \mapsto  \mathbb{R}^6$ is a function which converts a homogeneous transformation matrix to a spatial twist. The end-effector velocity $v_e$ can be substituted in (\ref{m:rrmc3}) to compute the corresponding joint velocities to reach the target. We use the Robotics Toolbox for Python \cite{rtb} for the PBS scheme and to calculate the forward and differential kinematics.

While this controller can reach arbitrary goals within the robot's workspace, it tends to result in poor manipulability performance across the trajectory. This reduces the robustness of the robot's behaviours and increases the chances of it hitting a singularity. Additionally, the robot's final pose tends to be ill-conditioned for the completion of a consecutive task. This renders it suboptimal with respect to the manipulability-based reaching task evaluated in this work.\\

\subsection{Training Details}

We leverage SAC as the underlying RL algorithm for BCF using the implementation provided by SpinningUp \cite{SpinningUp2018} with a discount factor $\gamma = 0.99$, learning rate of $1e-3$, replay buffer capacity of $1e6$ and batch size 100. We train an ensemble of 10 RL agents, each initialised randomly and updated using their own batch of experience sampled from a shared replay buffer. We found that this simple strategy together with the discrepancies in the target networks provided enough variation across the ensemble members over the course of training. \\ 

With the given experimental setup, we investigate to what extent BCF learns faster and safer than model-free RL alone, improves upon the given control prior, and its ability to safely deploy RL policies in the real world.

\begin{table*}[]
\caption{Summary of training results. BCF demonstrates accelerated learning, uncertainty-aware exploration, and achieves the highest improvement beyond the suboptimal control priors used in comparison to existing RLBP approaches.}
\resizebox{\textwidth}{!}{%
\label{summary}
\begin{tabular}{@{}lcccccc@{}}
\toprule
\multicolumn{1}{c}{\textbf{Property}} & \multicolumn{6}{c}{\textbf{Algorithm}} \\ \midrule \midrule
 & \multicolumn{1}{l}{\textbf{BCF (Ours)}} & \multicolumn{1}{l}{Residual RL} & \multicolumn{1}{l}{MCF} & \multicolumn{1}{l}{KL Regularised} & \multicolumn{1}{l}{CORE-RL} & \multicolumn{1}{l}{SAC} \\ \midrule
Accelerated Learning & \greencheck & \greencheck & \greencheck & \greencheck & \greencheck & \redcross \\
Uncertainty-Aware Exploration & \greencheck & \redcross & \redcross & \redcross & \greencheck & \redcross \\
\begin{tabular}[c]{@{}l@{}}Improvement from Control Prior - PointGoal Navigation \end{tabular} & \textbf{116}\% & 95\% & 109\% & 21\% & 42\% & - \\
\begin{tabular}[c]{@{}l@{}}Improvement from Control Prior - Maximum Manipulability Reacher\end{tabular} & \textbf{282}\% & 110\% & 229\% & - & - & - \\ \bottomrule
\end{tabular}
}
\end{table*}
\begin{figure*}[t]
  \centering
  \includegraphics[width=\textwidth]{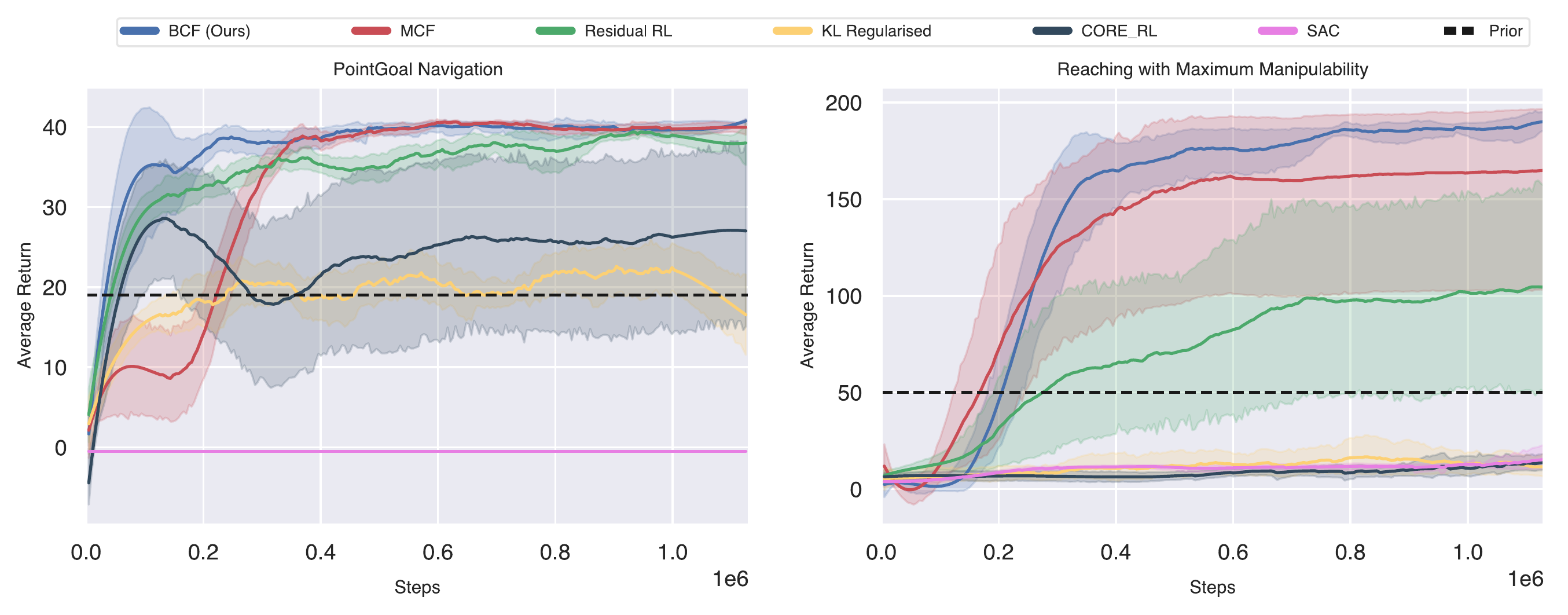}
  \caption{Learning curves of BCF and existing RLBP baselines for PointGoal Navigation and Reaching with Maximum Manipulability tasks. Note the faster convergence and lower variance across 10 seeds exhibited by our proposed approach.}
  \label{learn_curves}
\end{figure*}

\section{Evaluation of Training Performance}
\label{sec:training_performance}
We provide an evaluation of training performance when compared to four different RLBP baselines that have been proposed in related work. We additionally compare training curves for vanilla end-to-end trained policies and indicate the average performance of the control prior used. For all baselines, we utilise SAC as the underlying RL algorithm and train each system across 10 random seeds. We present the training curves for both tasks in Figure \ref{learn_curves} and a summary of the key characteristics demonstrated by each approach in Table \ref{summary}. We note here that the BCF training curves illustrate the evaluation of the performance of the standalone RL policy component in order to better depict how BCF can be used as a framework for accelerated RL policy training without being reliant on the presence of the control prior once the agent has converged.

\subsection{Baselines}

\begin{enumerate}
    \item \textit{Residual Reinforcement Learning:} Implementation of the residual reinforcement learning algorithm proposed by \cite{johannink2018residual}.
    \item \textit{KL Regularised RL:} Modified SAC algorithm which utilises a KL regularised objective towards a prior behaviour as proposed by \cite{pertsch2020accelerating}. This method utilises an auto temperature adjustment for the KL objective.
    \item \textit{CORE-RL:} Implementation of the TD-error-based exploration strategy to balance exploration between a control prior and the policy as proposed by \cite{cheng2019control}.
    \item \textit{MCF:} Our prior work that leverages a fixed gating function to switch between the control prior and policy over the course of training \cite{rana_mcf}.
    \item \textit{SAC:} Vanilla SAC algorithm using maximum entropy based Gaussian exploration \cite{haarnoja2019soft_}.
    \item \textit{Control Prior:} Classical controller based on the algorithms described in Section \ref{classical_controllers}.
\end{enumerate}


\subsection{Accelerated Learning}

\begin{figure*}[t]
  \centering
  \includegraphics[width=\textwidth]{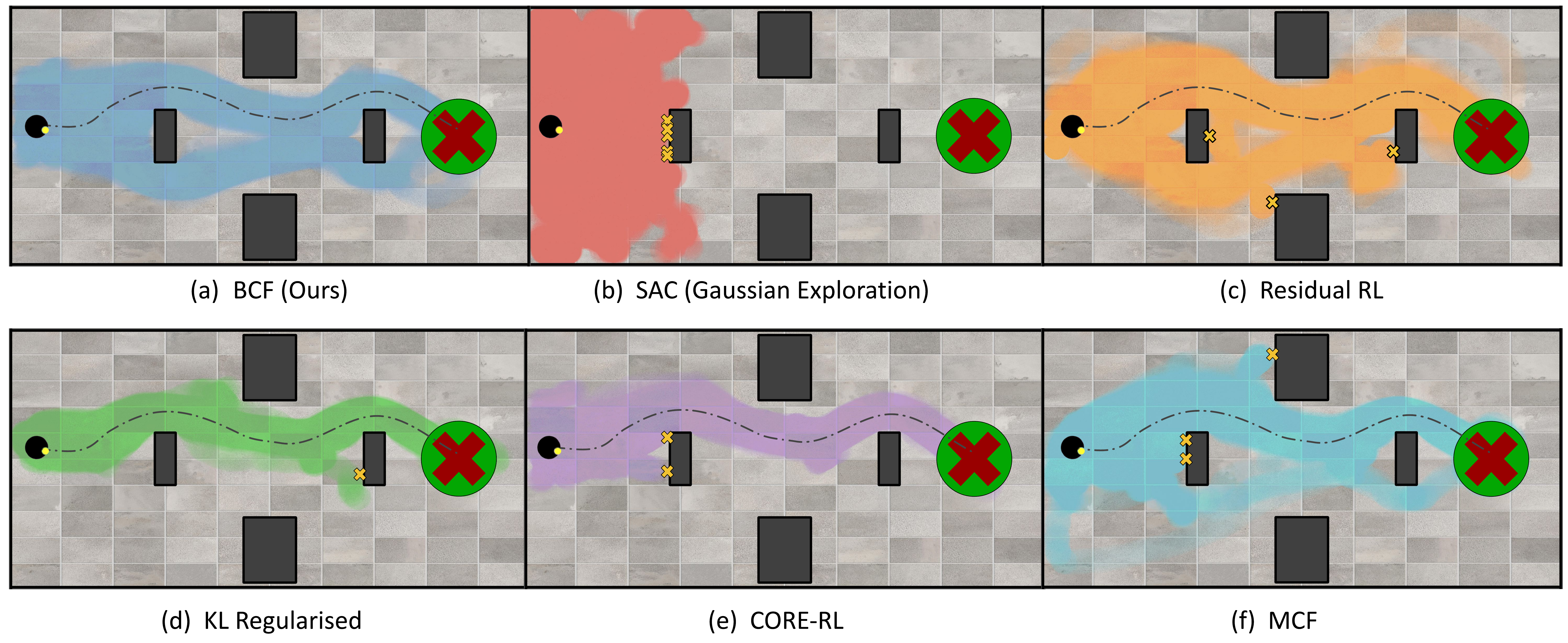}
  \caption{State space coverage during exploration. The dashed line illustrates the deterministic path taken by the control prior. Note how our formulation explores
  regions of the state space that are likely to lead to a meaningful progression to the goal, while still exploring a diverse region around the deterministic control prior for potential improvements. The yellow crosses indicate collisions with an obstacle that resulted in a failed episode.
  }
  \label{explore}
\end{figure*}

Across both tasks, BCF consistently demonstrates its ability to substantially accelerate training and achieve significantly higher final returns than the baselines. While all the RLBP baselines show the ability to accelerate learning when compared to vanilla SAC alone, it is also important to note their ability to improve beyond the control prior used. We quantify this improvement by computing the greatest change in performance attained by the approach as a fraction of the control prior's performance. The results are summarised in Table \ref{summary}.

For the navigation task, both KL-regularised RL and CORE-RL converge towards a final policy that exhibits suboptimal performance, while failing to learn at all in the reaching task. Residual RL and MCF both yield improvements beyond the control prior similar to that attained by BCF in the navigation task, attaining a 95\% and 109\% increase in performance respectively. However, in the reacher task, both approaches yield a substantially lower improvement when compared to that attained by BCF. We can speculate that this significant drop in performance between the two tasks is related to the performance gap between the control prior and the optimal policy for that task. For the Residual RL case, we can speculate that for highly suboptimal control priors, the residual's ability to express the required modifications to achieve the optimal behaviours is limited. The performance drop across MCF, KL regularised RL and CORE-RL may be related to the significant distribution mismatch between the control prior's behaviours and that of the current policy, where the current policy's behaviours are inadequately exploited. This can cause instabilities in the Q-value updates as seen in the offline RL setting \cite{kumar2019stabilizing,fujimoto2018addressing}. BCF on the other hand covers a broad distribution across both the control prior and its own behaviours, mitigating this phenomenon and achieving the best final performance.

To gain a better intuition into the success of BCF and the limitations of the existing RLBP approaches, we conducted a focused experiment in the navigation domain for a fixed start and goal location. Figure \ref{explore} shows the state-space coverage of the agent over the course of training for each of the RLBP approaches. The dashed line in the figure indicates the deterministic trajectory taken by the control prior, and the coloured regions indicate the states visited by the agent. The figure illustrates a key attribute across all RLBP approaches to bias the search space during exploration towards the most relevant regions for solving the task. This allows them to significantly accelerate training, particularly in the sparse, long-horizon reward setting. This is in stark contrast to the exhaustive exploration carried out by vanilla SAC, which hardly progresses beyond a third of the arena over the course of training, limiting its ability to learn at all.

As shown in Figure \ref{explore} (d) and \ref{explore} (e), the KL regularised and CORE-RL approaches both heavily constrain the behaviour of the policy towards that of the control prior used. This limits their ability to learn new behaviours. The stochastic nature of BCF on the other hand allows for a broader search space around the control prior's behaviours, allowing it to identify potentially optimal behaviours, while still biasing the agent towards the most relevant regions of the search space to accelerate learning.

\begin{figure}[t]
  \centering
  \includegraphics[width=0.5\textwidth]{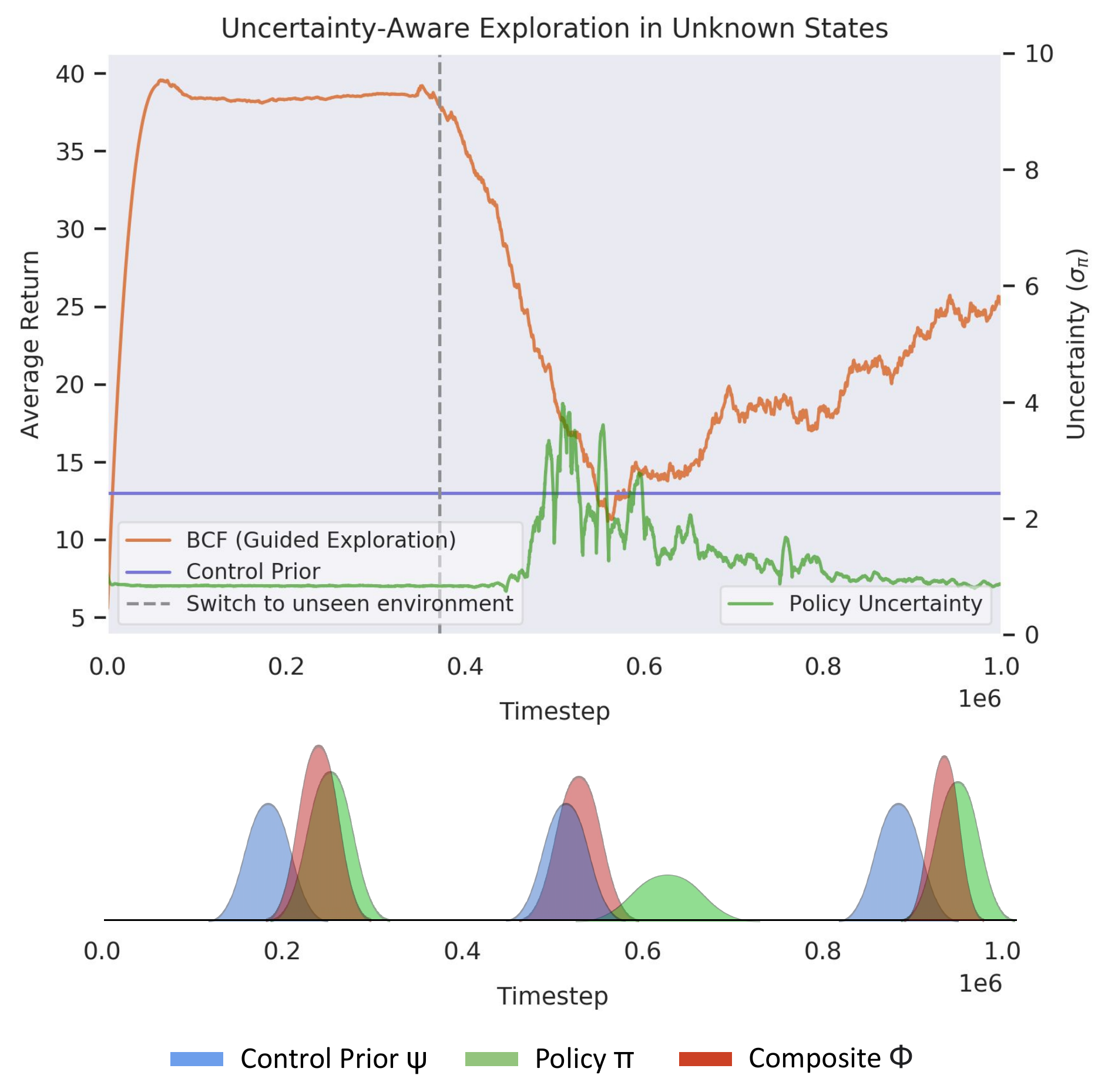}
  \caption{Uncertainty-aware exploration induced by BCF during training allows for safer behaviours when presented with unseen states. As the policy's uncertainty rises, BCF naturally relies more on the prior controller to guide exploration, while transitioning back to the policy once it gains more confidence. The accompanying plot at the bottom serves as a qualitative depiction of the BCF composition of the two distributions based on their relative uncertainty at the different stages of training.}
  \label{safety_aware}
\end{figure}

\subsection{Uncertainty-Aware Exploration}

We additionally investigate the ability of our uncertainty-aware exploration strategy to allow for safer exploratory behaviours when compared to existing RLBP approaches. This is a less explored area in existing RLBP literature that can have significant benefits as we gradually transition towards training these systems in the real world. Particularly in the case of robotics, we exploit the risk-aversity of the control priors developed and leverage these traits to allow for safer exploratory behaviours. As shown in Figure \ref{explore}, we additionally indicate obstacle collisions experienced by the agent that results in an overall failed episode during training. We mark the mean location for the collisions as yellow crosses in the figure. BCF completes training in this experiment without any collisions. We can attribute this to the uncertainty-aware formulation of the composite policy, which allows the risk-averse control prior to dominate control in states that the agent has never experienced before. This allows the agent to steer clear of these unsafe states throughout training, safely guiding the agent towards the goal. In contrast, all the baseline RLBP, while successfully constraining the search space, experience multiple collisions throughout training. It is important to note that while the KL regularised and CORE-RL approaches do exhibit a lower number of collisions on average, this comes at the expense of a significant drop to the overall optimality of the final policy, as they over-constrain exploration. BCF on the other hand is able to balance these two characteristics naturally.

In Figure \ref{safety_aware} we take a closer look at how the uncertainty-aware BCF formulation operates at the distributional output level when presented with unknown states during training. We explore its ability to balance the guided exploration provided by the control prior and the exploitation of the policy. We conduct the experiment in another focused PointGoal Navigation setting, where we expose the agent to different arenas over the course of training. We train the agent within the first arena until convergence and switch to a novel unknown environment after 390k steps, indicated by the dashed line in Figure \ref{safety_aware}. We indicate the performance of the agent over the course of training, as well as the empirical uncertainty of the policy. Directly below this graph, we illustrate a qualitative depiction of the distributional composition from BCF for the linear velocity component of the mobile robot at three key locations. As shown in the figure, once the agent has converged to the optimal policy, and the policy is highly certain of its surroundings, the policy component $\pi$ is predominantly governing the exploratory behaviours of the agent. Upon switching to the unknown environment, we see a significant increase in the uncertainty of the policy and the transition of the composite distribution $\phi$ towards the behaviours suggested by the control prior $\psi$. While we see a significant drop in performance towards that of the control prior, it is important to note that its risk-averse behaviours help guide exploration, allowing for safer exploratory behaviours than the highly uncertain black-box policy outputs. As the agent becomes more certain with its surrounding states, we see BCF transition control to the policy, allowing it to exploit its newly found behaviours. This additionally enables it to further explore surrounding state-action pairs allowing for improvements beyond the performance of the control prior.

\subsection{Impact of Control Prior Performance}
\begin{figure}[t]
  \centering
  \includegraphics[width=0.5\textwidth]{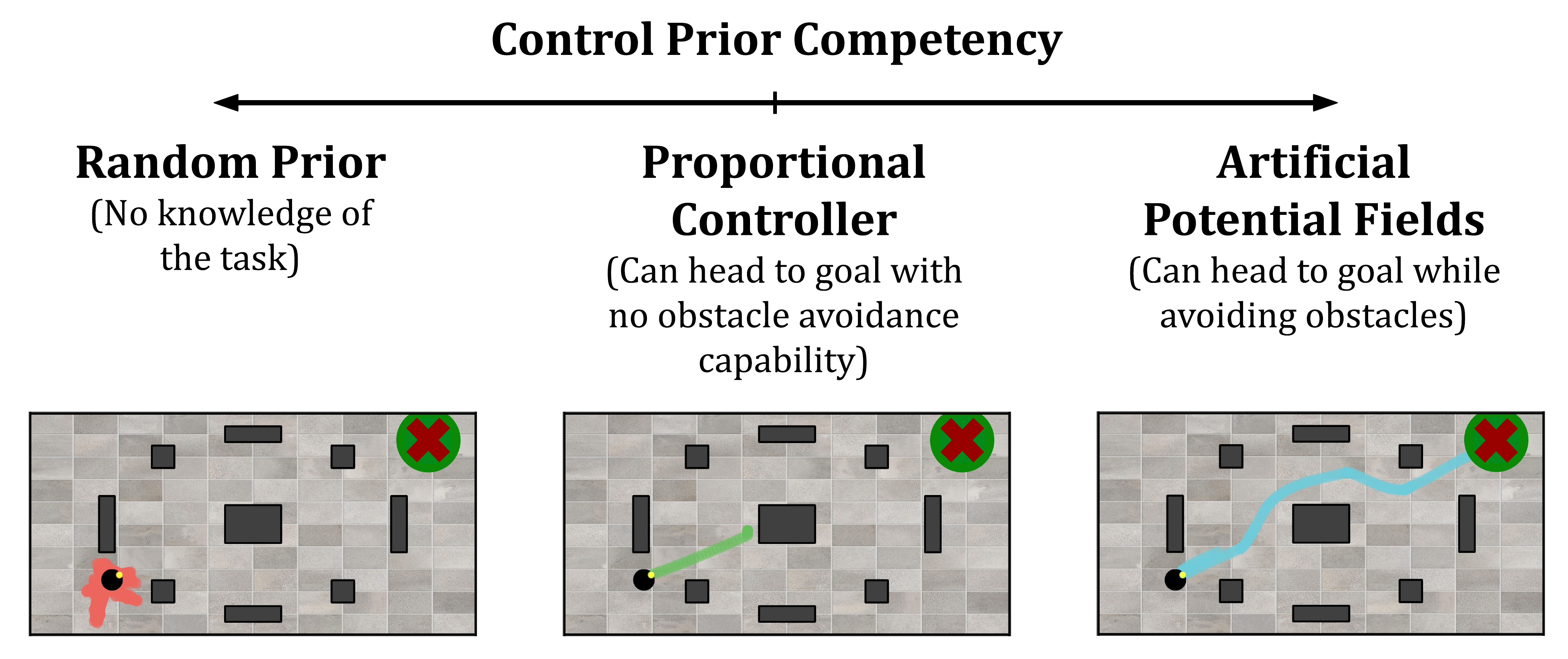}
  \caption{Control prior competency spectrum for the PointGoal Navigation task.}
  \label{prior_performance_spectrum}
\end{figure}
In this study, we analyse the impact that the competency of the control prior has on the overall training process. We seek to identify if too much structure from the control prior strongly biases the policy to a particular solution or if the control prior should solely serve as hints to guide the policy in a general direction. Figure \ref{prior_performance_spectrum} shows the range of control priors we evaluated with BCF in the PointGoal navigation environment. On the least competent end of the spectrum, we test a random prior which is a na\"ive controller that represents a system with no knowledge towards solving the task at hand. At the mid-range, we utilise a proportional (P) controller which is a simple controller based on the Euclidean distance to the goal. It provides basic structural knowledge towards reaching the goal, however, is incapable of avoiding any obstacles. For the more competent prior we utilise the standard APF controller used throughout this work that can head to the goal while avoiding obstacles. 

\begin{figure}[t]
  \centering
  \includegraphics[width=0.5\textwidth]{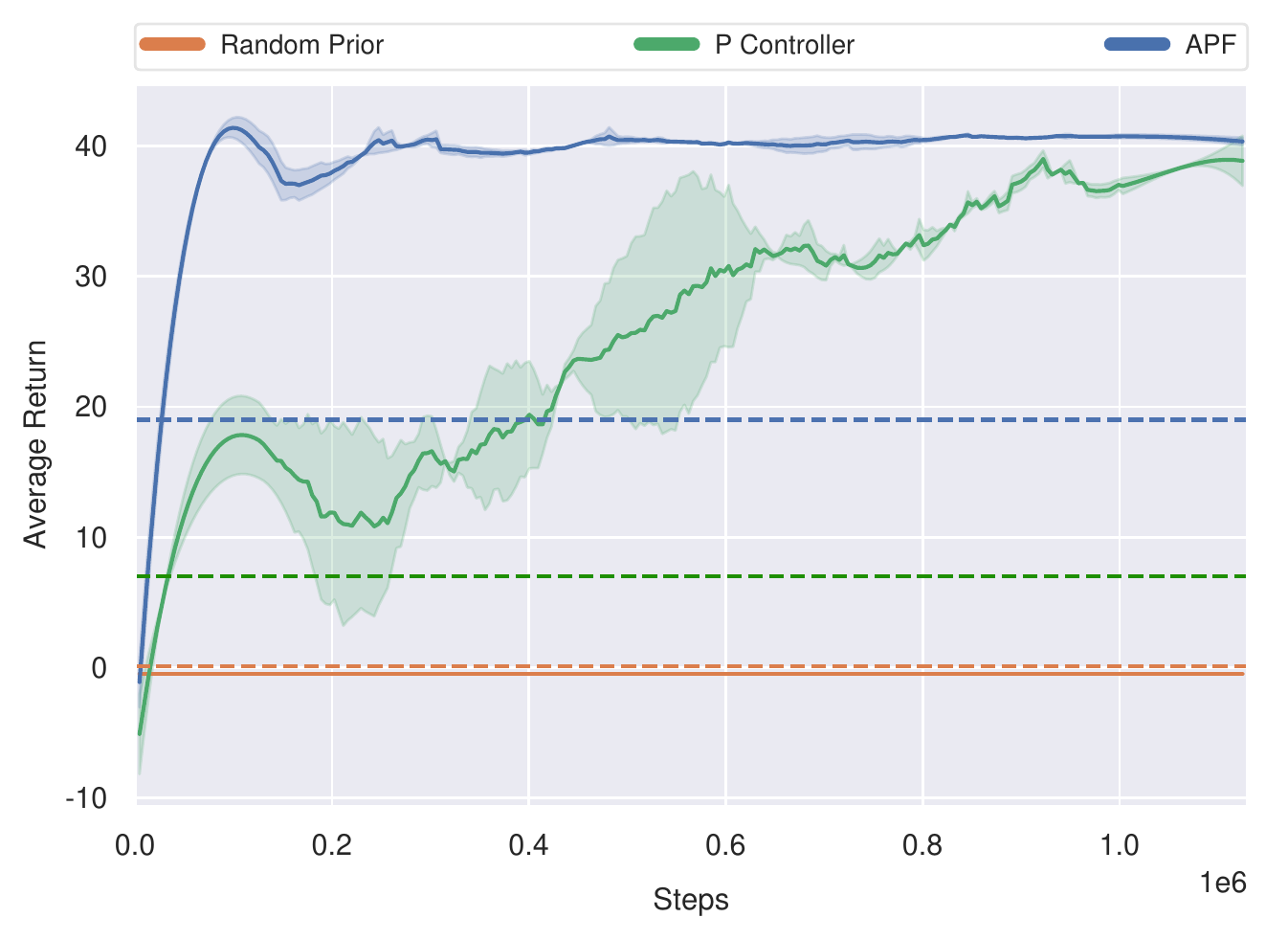}
  \caption{Training curves exploring the impact of the control prior's performance on the ability for BCF to accelerate learning in the PointGoal Navigation task. The corresponding dashed lines indicate the average return each control prior could achieve in the given environment.}
  \label{prior_performancw_ablation}
\end{figure}

The training curves for each of these controllers are given in Figure \ref{prior_performancw_ablation}. The random prior does not provide any additional benefit to exploration and the policy is incapable of learning within the given number of training steps. The controller exhibiting the most competency towards the task (APF) yielded the best performance in terms of sample efficiency and low variance. For the P-controller experiment, it is interesting to note that despite being a severely limited control prior, the agent still benefits from the exploration bias provided by it and is capable of attaining a higher-performing policy as training progresses. 
These results indicate that BCF is capable of leveraging a control prior to assist learning, and is not crippled from attaining a better policy regardless of the level of the initial performance of the control prior. This support the results observed for the reaching task as shown in Table \ref{summary}. It also suggests that control priors provide a useful form of positive bias to guide exploration as opposed to random exploration alone. It is important to note that despite their ability to accelerate learning, the competency of the control prior does play an important role when it comes to the safety of the agent. In such cases, the P-controller is not risk-averse and is prone to obstacle collisions. This makes it unsuitable when considering the application of BCF to safely train real-world systems.

\subsection{Impact of Control Prior Variance}

\label{sec:ablation_variance}
\begin{figure}[t]
  \centering
  \includegraphics[width=0.5\textwidth]{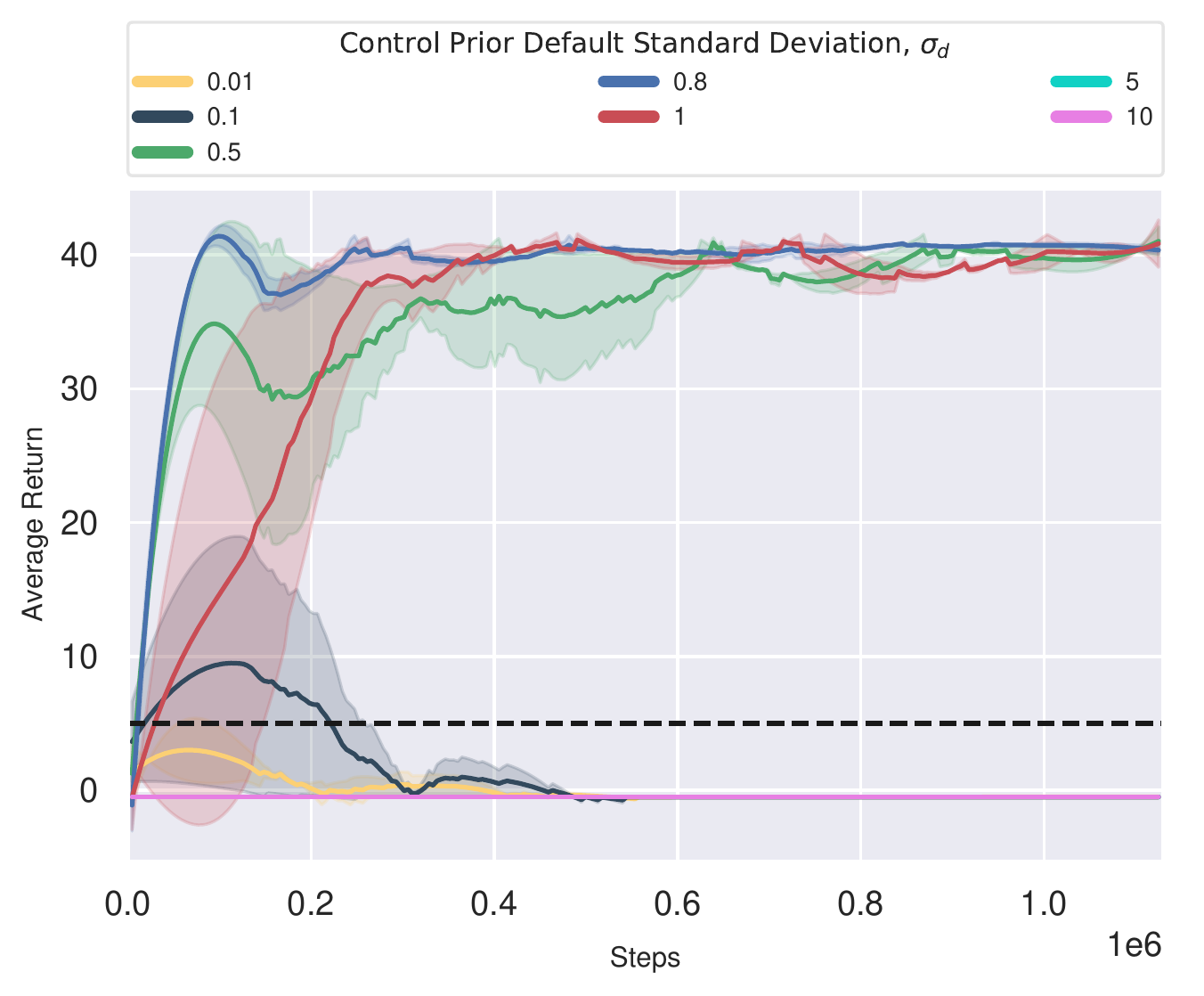}
  \caption{Learning curves for BCF using different default control prior standard deviations, $\sigma_d$. The dashed line indicates the average performance of the control prior in the environment.}
  \label{prior_sigma_curves}
\end{figure}

A key component of BCF is the distributional nature of the policy and the control prior. As most control priors are deterministic by nature, we approximate a Gaussian distribution by propagating any noise from the state inputs through to the action outputs using Monte Carlo sampling. We additionally set a default standard deviation $\sigma_d$ to allow for adequate exploration during training and to prevent the collapse of the distribution towards a deterministic value. In such a case the control prior would always dominate and the impact of the policy would be rendered ineffective. In practice, we found that this default value tended to dominate over the empirical distribution derived using MC sampling, given the inherent robustness to noise of the control priors used in this work. As the value of $\sigma_d$ is left as a hyper-parameter that needs to be carefully selected, we study its impact on the overall training performance of BCF by sweeping across a range of fixed standard deviation values. Note that the choice of $\sigma_d$ is dependent on the particular action type and its corresponding unit of measurement.

We conducted these experiments in the PointGoal navigation environment utilising the APF controller as the underlying control prior. The resulting learning curves are provided in Figure \ref{prior_sigma_curves}. The chosen standard deviation was fixed for both the linear and angular velocity components. With low standard deviation values, the agent fails to learn at all. In this setting, the control prior exhibits high confidence and hence strongly biases the composite sampling distribution towards its own behaviours. This limits the ability of the policy to exploit its own actions during exploration which results in it failing to learn. Such a setting is reminiscent of the offline RL setting, where the agent is solely trained on data not pertaining to its own behaviours, resulting in compounding errors due to overestimation bias. Standard deviation values set in the range of 0.5 to 1 exhibit the best performance. Such distributions provide a softer constraint on exploration allowing the agent to balance exploration and exploitation. Larger standard deviation values, greater than 1, resulted in the agent not learning at all. This is a result of the BCF formulation constantly rendering the policy as the more confident system and hence limiting the impact the control prior could have on the overall system. This is equivalent to the agent learning without any guidance from the control prior, similar to vanilla SAC.

\section{Evaluation of Deployed System}
\label{sec:evaluation_deployed}
An important motivation for this work is to leverage RL policies as a reliable strategy to control robots. In this section, we assess the ability of BCF to attain this during deployment and address the current limitations of both RL and classical robotics. As shown in Figure \ref{safety_aware}, BCF thrives in out-of-distribution states, a common occurrence when considering the sim-to-real setting. As opposed to a neural network-based policy catastrophically failing in these states, BCF has the ability to naturally transfer control to the risk-averse control prior which dominates control until a more suitable state is presented to the policy. In this setting, we gain the optimality of the learned policy in less uncertain states and the safety of the hand-crafted control prior otherwise. We thoroughly evaluate this control strategy for the two robotics tasks in both simulation and the real world. To better understand how the compositional nature of BCF can take advantage of the strengths of each individual component, we evaluate the individual components in isolation against our compositional BCF formulation. We provide details of the evaluated systems below.

\begin{enumerate} 
    \item \textit{Control Prior:} The deterministic classical controller derived using analytic methods.
    \item \textit{Policy Only:} SAC agent, trained to convergence using BCF, and deployed as a standalone policy.
    \item \textit{BCF:} Our proposed hybrid control strategy that combines uncertainty-aware action outputs from the control prior and trained RL policy.
\end{enumerate}

Note that all the policies used in this evaluation were trained to convergence in the simulation environments and directly deployed in the real world, without any fine-tuning. We provide a detailed account of each task in the following sections.

\subsection{PointGoal Navigation} In this experiment, we examine whether BCF could overcome the limitations of an existing reactive navigation controller, in this case, APF while leveraging this control prior to safely deal with out-of-distribution states that the policy could fail in. The APF controller exhibited suboptimal oscillatory behaviours particularly in between obstacles and tended to stagnate within local minima.

\subsubsection{Simulation Environment Evaluation}

For this task, we report the Success weighted by (normalised inverse) Path Length (SPL) metric proposed by \cite{anderson2018evaluation}. SPL weighs success by how efficiently the agent navigated to the goal relative to the shortest path. The metric requires a measure of the shortest path to the goal which we approximate using the path found by an A-Star search across a 2000 $\times$ 1000 grid. An episode is deemed successful when the robot arrives within 0.2m of the goal. The episode is timed out after 500 steps and is considered unsuccessful thereafter. We additionally report the average actuation time it takes for the agent to reach the goal.

As shown in Table \ref{tab_sim_nav}, we divide this evaluation across the known training environment and an unseen environment in order to better evaluate the impact of BCF. Across both settings, BCF attains superior performance when compared to the control prior. The lower actuation time indicates its ability to overcome the inefficient oscillatory motion while the higher SPL indicates its ability to attain a shorter path to the goal. More importantly, we note that BCF and the standalone policy attain the same performance in the known training environment. This is an important result that shows that BCF does not impact the optimality of the learned policy in states where the policy is confident to act. In the unseen environment, however, we see that BCF surpasses the performance of the standalone policy, given its ability to deal with out-of-distribution states reliably. We extend this evaluation to the sim-to-real setting, where we explore BCF as a reliable transfer strategy for an RL policy.

\begin{table}[t]
\caption{Evaluation of PointGoal Navigation in the Simulation Environment}
\label{tab_sim_nav}
\resizebox{0.49\textwidth}{!}{%
\begin{tabular}{@{}lcc
>{\columncolor[HTML]{C0C0C0}}c 
>{\columncolor[HTML]{C0C0C0}}c 
>{\columncolor[HTML]{FFFFFF}}l @{}}
\toprule
\multicolumn{1}{c}{} &
  \multicolumn{2}{c}{Training Environment} &
  \multicolumn{2}{c}{\cellcolor[HTML]{C0C0C0}Unseen Environment} &
   \\ \midrule
\textbf{Method} &
  \textbf{SPL} &
  \textbf{\begin{tabular}[c]{@{}c@{}}Actuation Time\\ (Steps)\end{tabular}} &
  \textbf{SPL} &
  \textbf{\begin{tabular}[c]{@{}c@{}}Actuation Time \\ (Steps)\end{tabular}} &
   \\ \midrule
 Control Prior                     &     0.299      &    462 $\pm$ 78.1         &   0.273      & 401 $\pm$ 149 &  \\ 
Policy Only                      &  0.958    &  141 $\pm$ 97.6      &  0.780    &     185 $\pm$ 179   &  \\
\textbf{BCF (Ours)} & \textbf{0.958} & \textbf{104 $\pm$ 11.9} & \textbf{0.909} & \textbf{149 $\pm$ 129} &  \\ \bottomrule
\end{tabular}
}
\end{table}

\subsubsection{Real-World Evaluation}

We utilise a GuiaBot mobile robot which is equipped with a 180$^\circ$ laser scanner, matching that used in the simulation environment. The velocity outputs from the policies are scaled to a maximum of \SI{0.25}{\meter/\second} before execution on the robot at a rate of 100 Hz. The system was deployed in a cluttered indoor office space that was previously mapped using the laser scanner. We utilise the ROS \texttt{AMCL} package to localise the robot within this map and extract the necessary state inputs for the policy network and control prior. Despite having a global map, the agent is only provided with global pose information with no additional information about its operational space. The environment also contained clutter which was unaccounted for in the mapping process. To enable large traversals through the office space, we utilise a global planner to generate target sub-goals, for our reactive agents to navigate towards. We do not report the SPL metric for the real robot experiments as we did not have access to an optimal path. We do, however, provide the distance travelled along each path and compare them to the distance travelled by a fine-tuned ROS \texttt{move\_base} controller. This controller is not necessarily the optimal solution but serves as a practical example of a commonly used controller on the Guiabot.

\begin{figure}[h]
  \centering
  \includegraphics[width=0.45\textwidth]{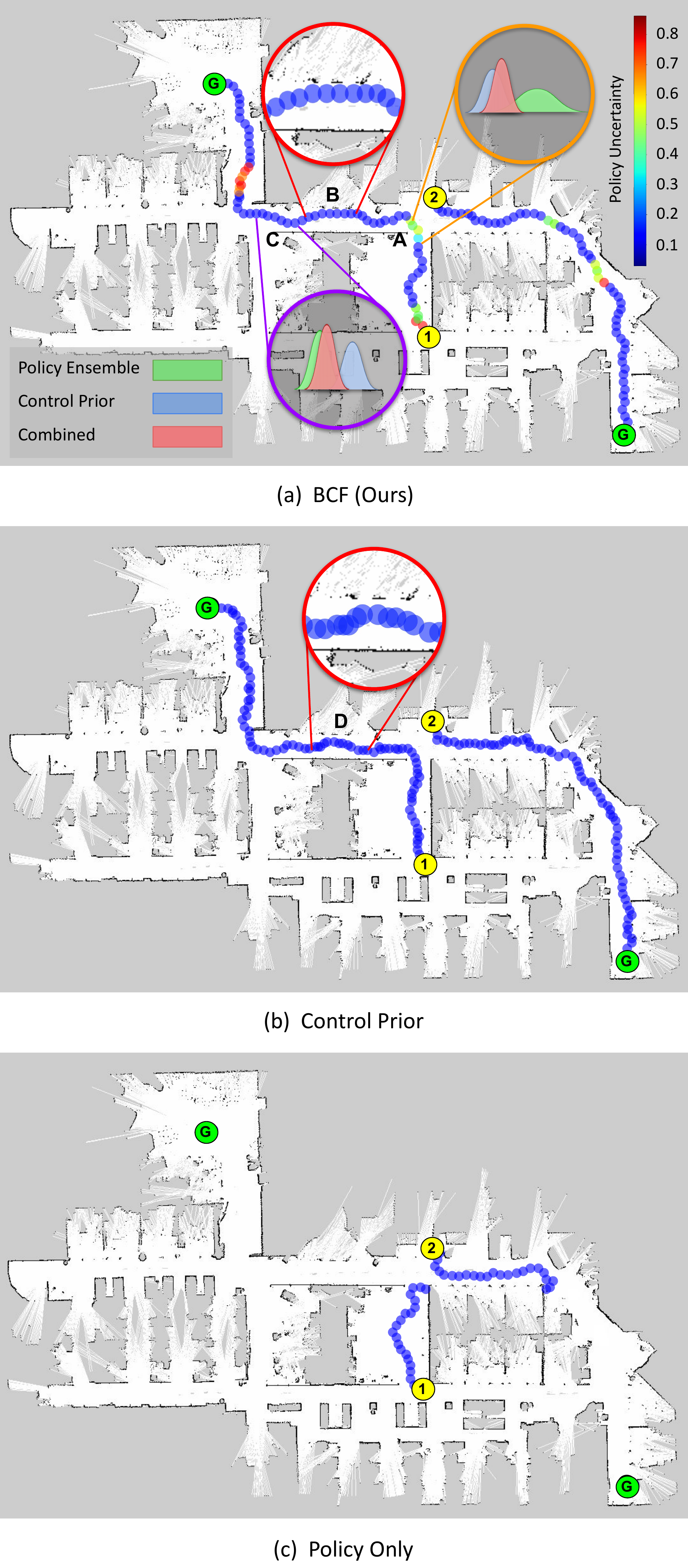}
  \caption{Trajectories taken by the real robot for the different start (orange) and goal locations in a cluttered office environment with long narrow corridors. The trajectory was considered unsuccessful if a collision occurred. The trajectory taken by BCF is colour coded to represent the uncertainty in the linear velocity of the trained policy. We illustrate the behaviour of the fused distributions at key areas along the trajectory. The symbols \textbf{1} and \textbf{2} indicate the start locations for each trajectory and \textbf{G} indicates the corresponding goal locations.}
  \label{maps}
  \vspace{-0.4cm}
\end{figure}

The evaluation was conducted on two different trajectories indicated as Trajectory 1 and 2 in Figure \ref{maps} and Table \ref{tab:real_robot}. Trajectory 1 consisted of lab space with multiple obstacles, tight turns, and dynamic human subjects along the trajectory, while Trajectory 2 consisted of narrow corridors never seen by the robot during training. We terminated a trajectory once a collision occurred and marked the run as a failed attempt. We summarise the results in Table \ref{tab:real_robot}. 

\begin{table}[t]
\caption{Evaluation of PointGoal Navigation in the Real-World}
\label{tab:real_robot}
\resizebox{0.5\textwidth}{!}{%
\begin{tabular}{@{}lcccc@{}}
\toprule
              & \multicolumn{2}{c}{\textbf{Trajectory 1}} & \multicolumn{2}{c}{\textbf{Trajectory 2}} \\ \midrule
\multicolumn{1}{c}{\textbf{Method}} &
  \textbf{\begin{tabular}[c]{@{}c@{}}Distance \\ Travelled \\ (meters)\end{tabular}} &
  \textbf{\begin{tabular}[c]{@{}c@{}}Actuation Time \\ (seconds)\end{tabular}} &
  \textbf{\begin{tabular}[c]{@{}c@{}}Distance \\ Travelled \\ (meters)\end{tabular}} &
  \textbf{\begin{tabular}[c]{@{}c@{}}Actuation Time \\ (seconds)\end{tabular}} \\ \midrule
Control Prior & 42.3                & 274                 & 35.3                & 277                 \\
Policy Only   & Fail                & Fail                & Fail                & Fail                \\
Move Base     & 62.6                & 263                 & 35.8                & 258                 \\
\textbf{BCF}  & \textbf{41.2}                & \textbf{135}                 & \textbf{30.4 }               & \textbf{117 }                \\ \bottomrule
\end{tabular}%
}
\end{table}

\label{sec:real_world}

Across both trajectories, the standalone policy failed to complete a trajectory without any collisions, exhibiting erratic reversing behaviours in out-of-distribution states. We can attribute this behaviour to its poor generalisation in such states, given the discrepancies in obstacle profiles seen during training in simulation and those encountered in the real world as shown in Figure \ref{robots} (a). The control prior was capable of completing all trajectories however required significantly long actuation times. We can attribute this to its inefficient oscillatory motion when moving through passageways and in between obstacles. BCF was successful across both trajectories exhibiting the lowest actuation times across all methods. This indicates its ability to exploit the optimal behaviours learned by the agent while ensuring it did not act erratically when presented with out-of-distribution states. It also demonstrates superior results when compared with the fine-tuned ROS \texttt{move\_base} controller.

To gain a better understanding of the reasons for BCF's success when compared to the control prior and the standalone policy acting in isolation, we examine the trajectories taken by these systems as shown in Figure \ref{maps}. The trajectory attained using BCF is colour-coded to illustrate the uncertainty of the policy's actions as given by the outputs of the ensemble. We draw the reader's attention to the region marked \textbf{A} which exhibits higher values of policy uncertainty. The composition of the respective distributions in this region is shown within the orange ring. Given the higher policy uncertainty at this point, the resulting composite distribution was biased more towards the control prior which displayed greater certainty, allowing the robot to progress beyond this point safely. We note here that this is the particular region that the standalone policy agent failed as shown in Figure \ref{maps} (c). The purple ring at region \textbf{C} illustrates a region of low policy uncertainty with the composite distribution biased closer towards the policy. Comparing the performance benefit over the control prior gained in such a case, we draw the reader's attention to regions \textbf{B} and \textbf{D} which show the path profile taken by the respective agents. The dense darker path shown by the control prior indicates regions of high oscillatory behaviour and significant time spent at a given location. On the other hand, we see that BCF does not exhibit this and attains a smoother trajectory which we can attribute to the learned policy component having higher precedence in these regions, stabilising the oscillatory effects of the control prior. This illustrates the ability of BCF to exploit the relative strengths of each component throughout the deployment stage.

\subsection{Maximum Manipulability Reacher}

We evaluate the ability of BCF to build upon the basic structure provided by an RRMC reaching controller for a 7 DoF arm robot in order to learn a more complex manipulability-maximising reaching controller. While RRMC provided the policy with the knowledge to reach a goal, the agent had to learn how to modify the individual joint velocities of each joint in order to maximise the manipulability of the controller.

\begin{figure*}[t]
  \centering
  \includegraphics[width=\textwidth]{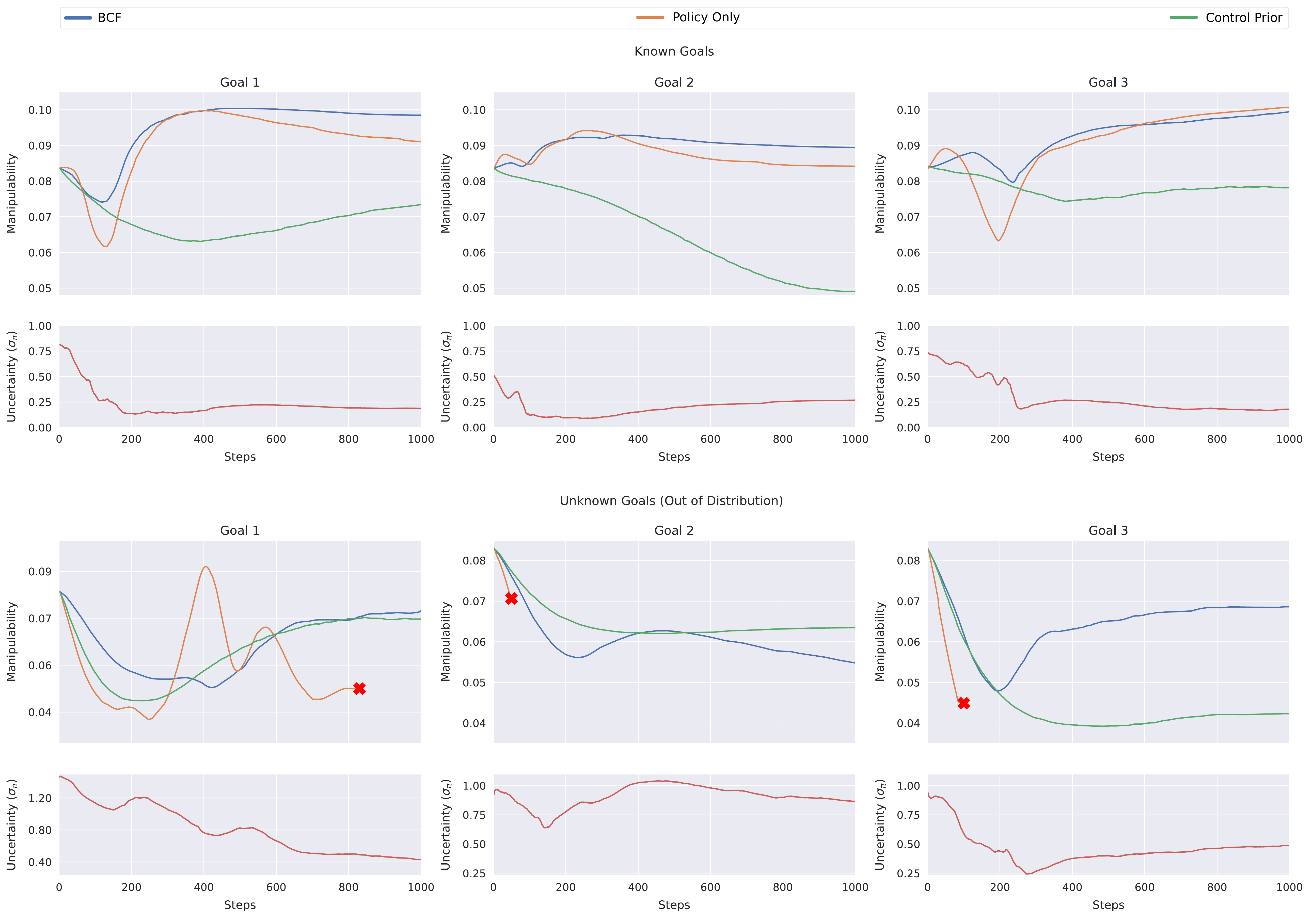}
  \caption{Manipulability and uncertainty curves for known and out-of-distribution goals for the reacher task, deployed on a real robot. The red cross indicates a failed trajectory.}
  \label{robot_manip_curves}
\end{figure*}

\begin{table}[t]
\caption{Evaluation of Maximum Manipulability Reacher in the Simulation Environment}
\label{tab_sim_man}
\resizebox{0.49\textwidth}{!}{%
\begin{tabular}{@{}lcc
>{\columncolor[HTML]{C0C0C0}}c 
>{\columncolor[HTML]{C0C0C0}}c 
>{\columncolor[HTML]{FFFFFF}}l @{}}
\toprule
\multicolumn{1}{c}{} &
  \multicolumn{2}{c}{Training Environment} &
  \multicolumn{2}{c}{\cellcolor[HTML]{C0C0C0}Unseen Environment} &
   \\ \midrule
\textbf{Method} &
  \textbf{\begin{tabular}[c]{@{}c@{}}Average\\ Manipulability\end{tabular}} &
  \textbf{\begin{tabular}[c]{@{}c@{}}Success\\ Rate\end{tabular}} &
  \textbf{\begin{tabular}[c]{@{}c@{}}Average\\ Manipulability\end{tabular}} &
  \textbf{\begin{tabular}[c]{@{}c@{}}Success\\ Rate\end{tabular}} &
   \\ \midrule
 Control Prior            &  0.0633  $\pm$  0.0126    &    100\%      &   0.0628 $\pm$ 0.0197     &  100\% &  \\ 
Policy Only               & 0.0949 $\pm$ 0.00476     &    100\%  & 0.0810 $\pm$ 0.0296    &   80\%    &  \\
\textbf{BCF (Ours)} & \textbf{0.0972 $\pm$ 0.00593} & \textbf{100\%} & \textbf{0.0924 $\pm$ 0.0161} & \textbf{100\%} &  \\ \bottomrule
\end{tabular}
}
\end{table}

\subsubsection{Simulation Environment Evaluation} For this task, we report the average manipulability across an entire trajectory and the success rate of the agent out of 10 trials. Table \ref{tab_sim_man} summarises the results. The robot was trained with a subset of goals randomly sampled from the positive x-axis region of its workspace frame as shown in Figure \ref{robots}. We classify goal states sampled from outside this region as out-of-distribution states during evaluation. Similar to the navigation task, BCF attains the best performance in both settings, improving the manipulability of the control prior by 34.9$\%$ without any failure cases. While the standalone policy agent successfully attained optimal behaviours in goal states within the training distribution, it exhibited erratic and unsafe behaviours when presented with out-of-distribution goal states. In these cases, the robot was seen to constantly crash into the countertop or hit its joint limits.

\subsubsection{Real-World Evaluation}

To ensure that the simulation-trained policies could be transferred directly to a real robot, we matched the coordinate frames of the PyRep simulator with the real Franka Emika Panda robot setup shown in Figure \ref{robots}. The state and action space were matched with that used in the training environment, with the actions all scaled down to a maximum of \SI{1.74}{\radian\per\second} before publishing them to the robot at a rate of 100 Hz. 

Table \ref{tab:real_arm} shows the results obtained when evaluating the agent on a random set of goals sampled from the robot's entire workspace. In all cases, BCF attains the highest manipulability and success rate surpassing both the control prior and standalone policy illustrating its ability to deal with higher dimensional action spaces. We take a closer look at individual trajectories across known and out-of-distribution goal states in order to better understand how BCF attains successful trajectories when compared to a standalone RL policy. Figure \ref{robot_manip_curves} shows the manipulability curves of the robot across these two sets, for three different goals sampled from each region. For each goal, we additionally indicate the performance of the control prior and standalone policy, together with a separate plot of the ensemble uncertainty estimate of the RL policy component in BCF across the trajectory indicated by the red curves.

\begin{table}[t]
\caption{Evaluation of Maximum Manipulability Reacher in the Real-World}
\label{tab:real_arm}
\resizebox{0.5\textwidth}{!}{%
\begin{tabular}{lccc}
\toprule
\multicolumn{1}{c}{\textbf{Method}} &
  \textbf{\begin{tabular}[c]{@{}c@{}}Average \\ Manipulability\end{tabular}} &
  \textbf{\begin{tabular}[c]{@{}c@{}}Average Final \\ Manipulability\end{tabular}} &
  \textbf{Success Rate} \\ \midrule
Control Prior &     0.0629$\pm$0.00926  &     0.0658$\pm$0.0165      &  98.2\%     \\
Policy Only   &     0.0803$\pm$0.00514      &   0.07812$\pm$0.0150        &  78.6\%         \\
\textbf{BCF}  & \textbf{0.0836$\pm$0.0156} & \textbf{0.0889$\pm$0.0177} & \textbf{98.2\%} \\ \bottomrule
\end{tabular}%
}
\end{table}

In the case of the known goals, BCF and the standalone RL policy both attain similar performances, maximising the manipulability of the agent across the trajectory. This is in stark contrast to the control prior which exhibits significantly poor performance across the trajectory. Note here that while the control prior exhibits poor performance with regard to manipulability, it is still successful in completing the reaching task at hand without any failures. It is interesting to note the high uncertainty of the ensemble at the start of a trajectory which quickly drops to a significantly lower value. The high uncertainty could be a result of the multiple possible trajectories that the robot could take at the start, which quickly narrows down once the robot begins to move. Note that once the policy ensemble exhibits a lower uncertainty, the performance of BCF closely resembles that of the standalone RL policy, indicating that BCF does not significantly impact the optimality of the learned behaviours.

When evaluating the agents on out-of-distribution goals, BCF plays an important role in ensuring that the robot can successfully and safely complete the task. Note the higher levels of uncertainty across these trajectories when compared to the case of the known goals. In all these cases, the standalone policy fails to successfully complete a trajectory, frequently self-colliding or exhibiting random erratic behaviours. We indicate these failed trajectories with a red cross in Figure \ref{robot_manip_curves}. BCF is seen to closely follow the behaviours of the control prior in states of high uncertainty, averting it from such catastrophic failures. While the composite control strategy works well to ensure the safety of the robot, the higher reliance of the system on the control prior results in suboptimal behaviour with regard to manipulability. The trade-off between task optimality versus the safety of the robot is an interesting dilemma that BCF attempts to balance naturally. The fixed standard deviation chosen for the control prior could serve as a tuning parameter to allow the user to better control this trade-off at deployment. A smaller standard deviation would bias the resulting distribution more strongly towards the control prior yielding more conservative and suboptimal actions; whereas a larger standard deviation would allow the agent to exploit the learned behaviours for potentially better performance, however at the expense of the robot's safety when in out-of-distribution states. We leave the further exploration of this idea to future work.

We provide videos illustrating the behaviours of the real robot on our project page \footnote{\url{https://krishanrana.github.io/bcf}}.

\section{Limitations}
\label{sec:limitations}
While we demonstrate BCF's ability to address some of the critical challenges faced by RL for robotics in both the training and deployment settings, there are a few limitations of our method that future work should consider. Firstly, BCF was primarily evaluated on tasks that involve free space motion which do not require any complex interaction between the robot and the environment. While BCF should naturally extend to these scenarios given the presence of a competent handcrafted control prior a thorough evaluation of the approach on these tasks is still to be made.

\begin{figure}[t]
  \centering
  \includegraphics[width=0.5\textwidth]{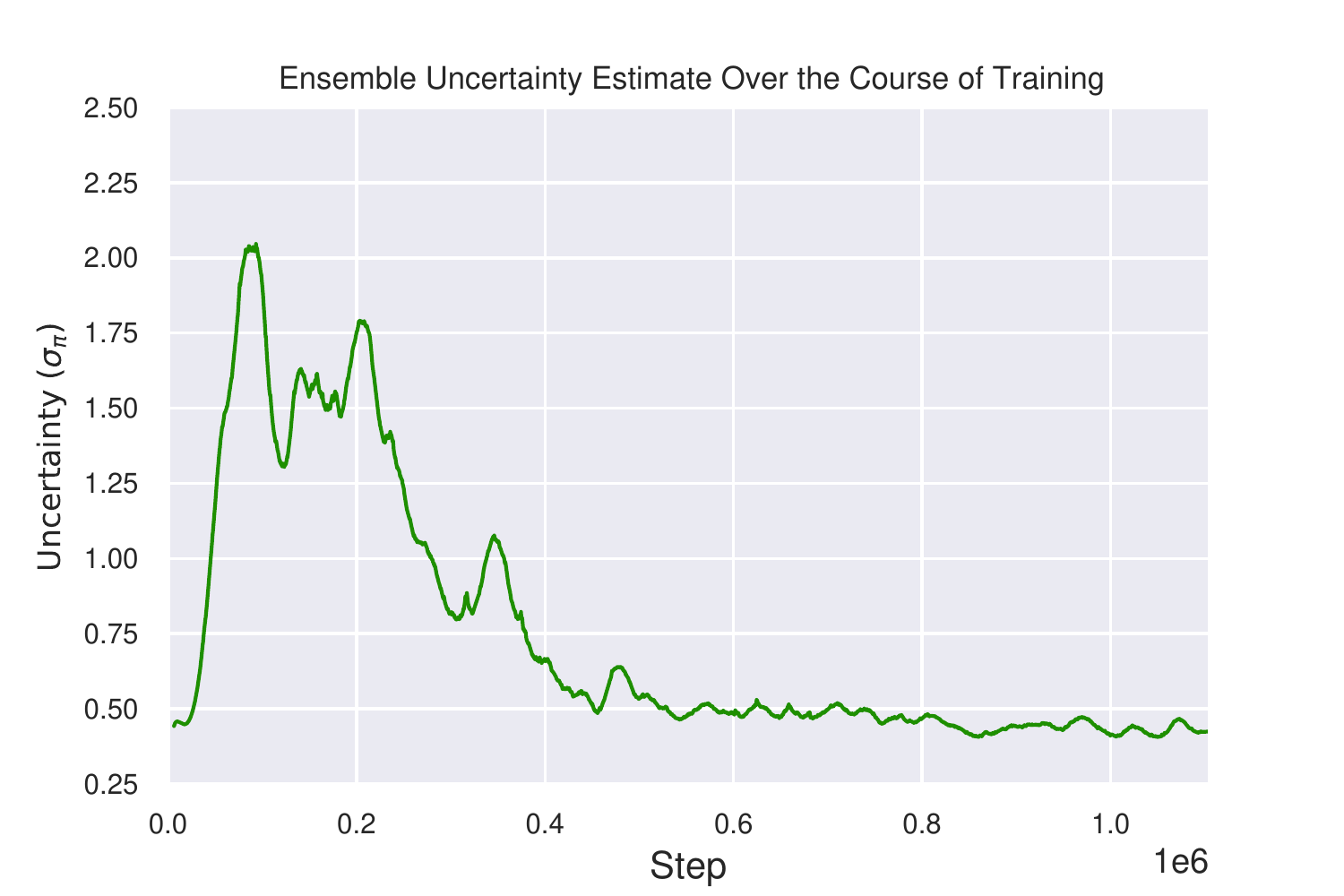}
  \caption{Progression of ensemble epistemic uncertainty estimates over the course of training. Note how it estimates relatively low uncertainties at the early stages of training before the ensemble members diverge.}
  \label{ep_uncertainty}
\end{figure}

Secondly, the uncertainty estimation technique using an ensemble of policies occasionally tended to underestimate the variance of the resulting distribution during the early stages of training. This was due to all the networks in the ensemble producing small values close to zero \cite{Osband2017DeepEV}. This poses a limit on the guidance that the control prior can provide to the agent within the BCF formulation. We did find in practice however that the ensemble members were able to quickly diversify after only a few gradient updates. We illustrate this phenomenon in the ensemble uncertainty estimate over the course of training in Figure \ref{ep_uncertainty}. One potential avenue future work could explore is the use of Randomized Prior Functions (RPFs) suggested by \cite{Osband2017DeepEV} who show that we can enforce this diversity from the start of training by pairing each network with a fixed, untrained network, which adds its predictions to the output. RPFs ensure a high variance at regions of the input space that are not explored, and a lower variance when the trained networks have learned to compensate for their respective priors and converge to the same output.

Another limitation we identified in this work is that the BCF formulation reduces the individual distribution outputs from the control prior and RL policy to a single univariate Gaussian for each actuator which limits the overall expressivity of the agent when presented with complex multi-modal scenarios. In such a case, each controller would be highly confident with opposing action distributions resulting in BCF averaging the two outputs which could produce highly unsuitable or unsafe behaviours. This calls for a more expressive representation of the resulting composition such as a Gaussian Mixture Modal.

Finally, our particular formulation is restricted to leveraging only a single control prior for guiding an RL agent. This could be restrictive in cases where multiple control priors exist, each exhibiting strengths that we would like the RL agent to capture. Future work could explore how we could incorporate these systems into a single learning framework for accelerated and safe learning. One avenue would be to incorporate the systems within a hierarchical RL setting. In such a case the high-level agent would be pre-trained to learn an effective exploration-exploitation arbitration strategy to govern the choice among the different controllers and policies based on their uncertainty over actions and a given state.

\section{Conclusion}
\label{sec:conclusion}

Building on the large body of work already developed by the robotics community can greatly help accelerate the use of RL-based systems, allowing us to develop better controllers for robots as they move towards solving more complex tasks. The ideas presented in this paper demonstrate a strategy that closely couples traditional controllers with learned systems, exploiting the strengths of each approach in order to attain more reliable and robust behaviours. We see this as a promising step towards bringing reinforcement learning to real-world robotics.

Our Bayesian Controller Fusion (BCF) approach combines uncertainty-aware outputs from the two control modalities. In doing this, we show that we not only accelerate training but additionally learn a final policy that can substantially improve beyond the performance of the handcrafted controller, regardless of its degree of sub-optimality. We show results across both a navigation and reaching task where BCF attains a final policy exhibiting a 116\% and 282\% improvement beyond the initial performance of the control prior used respectively, substantially higher than that attained by existing approaches. More importantly, we show that our approach can exploit the risk-aversity provided by these classical controllers to allow for safe exploratory behaviours when presented with unknown states. 

At deployment, we show that forming a hybrid controller with BCF allows us to exploit the respective strengths of each controller, enabling the reliable performance of RL policies in the real world. Across two real-world tasks for navigation and reaching, we show that BCF can safely deal with out-of-distribution states in the sim-to-real setting, succeeding where a typical standalone policy would fail, while attaining the optimality of the learned behaviours in known states. In the navigation domain, we overcome the inefficient oscillatory motion of an existing reactive navigation controller, decreasing the overall actuation time during real-world navigation runs by 50.7\%. For the reaching task, we show that our hybrid controller achieves the highest success rate, and improves the manipulability of an existing reaching controller by 34.9\%, a system typically difficult to attain using analytical approaches.

\section{Acknowledgements}
We acknowledge continued support from the Queensland University of Technology (QUT) through the Centre for Robotics. This research was supported by the Australian Research Council Centre of Excellence for Robotic Vision (project number CE140100016). This work was partially supported by an Australian Research Council Discovery Project (project number DP220102398). The authors would like to thank Jake Bruce, Robert Lee, Mingda Xu, Dimity Miller, Thomas Coppin and Jordan Erskine for their valuable and insightful discussions towards this contribution.

\bibliography{IEEEabrv,references.bib,manual.bib}


\end{document}